%% file: acl_latex.tex
\pdfoutput=1

\documentclass[11pt]{article}

\usepackage[nohyperref]{acl}

\usepackage{times}
\usepackage{latexsym}
\usepackage{tabularx}
\usepackage{multirow}
\usepackage{booktabs}
\usepackage{makecell}
\usepackage[T1]{fontenc}
\usepackage{amsmath}
\usepackage[utf8]{inputenc}
\usepackage{microtype}
\usepackage{inconsolata}
\usepackage{graphicx}
\usepackage{enumitem}
\usepackage{xcolor}
\usepackage{colortbl}

\usepackage{tikz}
\usetikzlibrary{arrows.meta, positioning, shapes.geometric, fit, calc}
\usepackage[most]{tcolorbox}

\title{Beyond Individual Personas: Aligning Synthetic Dialogue to Population-Level Behavior Distributions}

\author{
 \textbf{Xinyi Liu\textsuperscript{1,2}}\thanks{Work done during an internship at Amazon.},
 \textbf{Rinat Khaziev\textsuperscript{1}},
 \textbf{Hooshang Nayyeri\textsuperscript{1}},
 \textbf{Emine Yilmaz\textsuperscript{1,3}},
 \textbf{Charith Peris\textsuperscript{1}},
 \textbf{Hari Thadakamalla\textsuperscript{1}}
\\
\\
 \textsuperscript{1}Amazon,
 \textsuperscript{2}University of Illinois Urbana--Champaign,
 \textsuperscript{3}University College London
\\
 \small\texttt{\{xinyiapr, rinatk, hooshang, perisc, thadakah\}@amazon.com}
\\
 \small\texttt{eminey@amazon.co.uk}
\\
 \small\texttt{Correspondence: liu323@illinois.edu}
}

\begin{document}
\maketitle

\begin{abstract}
Synthetic dialogue corpora are increasingly used as proxies for target dialogue data, yet persona-grounded generators optimize individual conversations rather than corpus composition, yielding locally plausible dialogues with distorted population-level behavior mixes. We introduce \textbf{GroupPersona}, a framework that aligns synthetic dialogue corpora to the behavior distribution of a reference corpus. GroupPersona turns population statistics into generation controls: it separates each dialogue's core behavioral signature from predictable side effects, and uses the resulting behavioral groups to condition user agents on the interaction patterns that define the reference population.

We evaluate GroupPersona on four corpora crossing two dialogue sources, assistant-style and Reddit-derived, with two construction variants: structure-preserving and variation-enhanced.
GroupPersona lowers Jensen--Shannon divergence between synthetic and reference distributions over 12 behavior attributes from 0.234 to 0.177 relative to the strongest average baseline, a 24.4\% reduction, and is best or tied-best on all four corpora while preserving structural alignment.
It also achieves the closest calibration to reference-conversation quality scores, reducing mean absolute deviation from the reference-conversation profile to 0.63 versus 0.91 for the next-best baseline.
\end{abstract}

\input{sections/02_intro}

\input{sections/03_related}

\input{sections/04_formulation}

\input{sections/05_method}

\input{sections/06_dataset}

\input{sections/07_experiments}

\input{sections/08_limitations}

\bibliography{custom}

\input{sections/09_appendix}

\end{document}

%% file: sections/02_intro.tex
\section{Introduction}
\label{sec:intro}

\begin{figure*}[t!]
\centering
\includegraphics[width=0.96\textwidth]{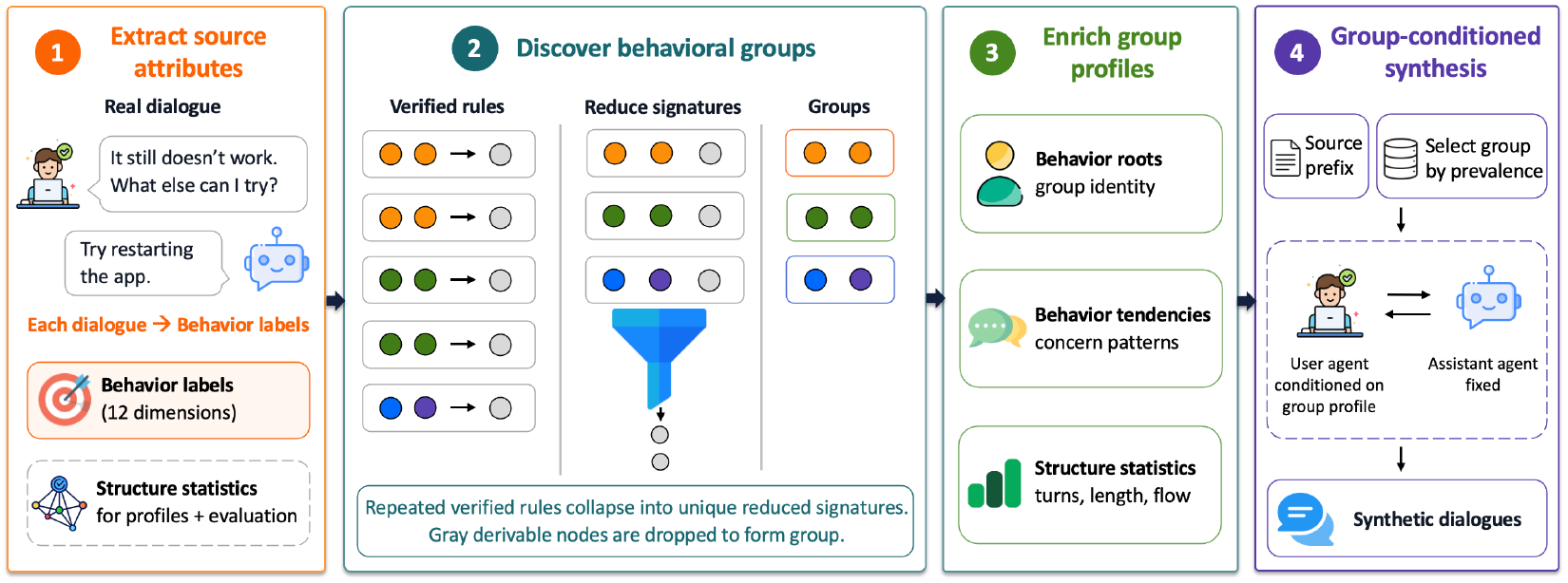}
\caption{\textbf{GroupPersona overview.} The pipeline extracts source attributes, removes rule-derivable behavior labels to form behavioral groups, enriches group profiles, and uses prevalence-weighted groups for dialogue generation.}
\label{fig:framework}
\end{figure*}

Persona-grounded dialogue synthesis has become a common way to create realistic conversational data. By assigning a persona or user profile to a generator, prior work can produce dialogues that are fluent, coherent, and locally consistent with a user description~\citep{zhang2018personalizing, jandaghi2024faithful, cheng2024autopal, wang2025know}. These synthetic dialogues are increasingly used beyond individual examples: they expand scarce training data, support user-simulator evaluation, stress-test assistants, and build personalization benchmarks~\citep{suresh2025diasynth, li2025mads, dou-etal-2025-simulatorarena, zhao2025personalens}. In these settings, generated dialogues function as corpus-level proxies for target user populations.

This corpus-level use exposes a failure mode that per-dialogue persona checks cannot capture. A generator may produce dialogues that are each fluent and persona-consistent, yet still generate the wrong mix of user behaviors. It may overproduce exploratory users, underproduce brief task-oriented users, or miss users who persist after system failures. The problem is therefore not local implausibility, but population-level mismatch: the generated corpus no longer reflects the behavior distribution of the reference corpus. We call this failure mode \emph{behavior-distribution misalignment}.

This failure mode also changes the evaluation target. When synthetic dialogues are used as corpus-level proxies, higher dialogue-quality scores are not automatically better: an over-polished synthetic corpus can score above reference conversations while moving away from the reference population. We therefore use behavioral distribution divergence as the primary target, check structural divergence to rule out shifts in basic dialogue form, and use LLM-as-Judge quality scores only as a secondary diagnostic: synthetic conversations should stay close to the reference-conversation quality range, not simply score higher (\S\ref{sec:exp-q4}; full rubric in App.~\ref{app:quality_calibration}).

The central challenge is to turn corpus-level behavior statistics into generation-time control. Such a control unit must be frequent enough to estimate from data, expressive enough to capture recurring behavior mixtures, and interpretable enough to guide an LLM user agent. Existing alternatives satisfy these requirements only partially. Individual personas provide local conditioning but do not specify corpus composition~\citep{zhang2018personalizing, jandaghi2024faithful, cheng2024autopal, wang2025know}. Exact behavior tuples make composition explicit but fragment the data into sparse joint cases, a standard issue in categorical pattern spaces~\citep{agrawal1993mining, agrawal1994fast, han2007frequent}. Latent clusters reduce sparsity, but their identifiers are difficult to translate into generation instructions.

We introduce \textbf{GroupPersona}, a framework that turns population behavior patterns into group-conditioned synthesis controls. Figure~\ref{fig:framework} gives the overview. GroupPersona separates source attributes into behavior attributes for group discovery and structural statistics for profile enrichment and evaluation. This design decouples behavioral identity from structural realization: groups are defined by recurring user-interaction patterns, while turn count and utterance length guide dialogue shape without determining group membership. A second design principle is that behavior attributes are not equally informative. Some define the core interaction type, while others are predictable side effects. For example, command-style task requests often imply short user utterances. Treating these side effects as independent clustering dimensions inflates sparsity and lets derivable attributes distort group structure.

To separate interaction types from predictable side effects, GroupPersona treats each dialogue as a set of behavior attribute--value pairs and mines association rules that predict one behavior label from a small set of co-occurring labels. An LLM judge filters statistically strong but behaviorally uninformative rules. Verified rules mark predictable labels as consequences, so clustering is performed on the remaining core behavior labels.

These groups are converted into user-agent profiles for synthesis. Each profile contains behavior roots that define the group, behavior tendencies that summarize common non-root patterns, and structural statistics that guide dialogue shape. During synthesis, GroupPersona selects behavioral groups according to their reference-corpus prevalence and provides the corresponding group profiles to the user agent, while the assistant agent remains fixed and group-blind. This design makes the user side carry the population signal while keeping assistant behavior constant across methods. The generated corpus is then evaluated by comparing its behavioral and structural distributions against the reference corpus.

We make three contributions:
\begin{itemize}[leftmargin=*, itemsep=0pt, topsep=1pt, parsep=0pt, partopsep=0pt]
\item \textbf{Population alignment target.} We show that persona-consistent dialogues can still yield a behaviorally misaligned corpus, and define synthetic dialogue evaluation as matching reference behavior distributions while preserving structure.
\item \textbf{Group-conditioned control.} 
We propose \textbf{GroupPersona}, which makes population alignment actionable by distilling each dialogue to its core behavioral signature and converting the resulting groups into prevalence-weighted controls for user-agent synthesis.
\item \textbf{Alignment gains.} GroupPersona improves behavioral alignment by $24.4\%$, preserves structure, and reduces the average quality-score gap to reference conversations to $0.63$ versus $0.91$ for the next-best baseline.
\end{itemize}

%% file: sections/03_related.tex
\section{Related Work}
\label{sec:related}

\paragraph{Persona-grounded dialogue synthesis and user simulation.}
Personalized dialogue generation conditions models on explicit personas, user profiles, or interaction histories to improve local consistency with an individual speaker~\citep{li2016persona, mazare2018training, zhang2018personalizing, wolf2019transfertransfo}. Recent work strengthens this paradigm through concept expansion, faithful persona-based data generation, user adaptation, implicit profiles, long-term histories, and customizable role-playing agents~\citep{kim2023concept, jandaghi2024faithful, cheng2024autopal, wang2025know, li2025toward, yang2025crafting}. LLM-generated dialogues are also widely used for low-resource data generation, multi-agent self-play, user-simulator evaluation, and personalization benchmarks~\citep{suresh2025diasynth, li2025mads, dou-etal-2025-simulatorarena, zhao2025personalens}. Together, these works optimize or evaluate synthetic dialogues primarily at the conversation level.
GroupPersona instead targets corpus-level alignment: whether generated dialogues preserve the behavior distribution of the reference population.

\paragraph{Corpus-level fidelity and controllable behavior groups.}
Distributional evaluation shows that plausible individual generations can still mismatch target data in aggregate properties~\citep{hashimoto2018unifying, li2016diversity, holtzman2020nucleus, welleck2020neural}, while dialogue evaluation motivates behavior-aware criteria beyond scalar quality scores~\citep{mehri2020usr, finch2020unified}. 
GroupPersona focuses on the missing link between measurement and generation: turning reference behavior distributions into interpretable controls for synthesis. 
To build these controls, we combine association-rule mining for transparent behavior dependencies~\citep{agrawal1993mining, agrawal1994fast, han2007frequent} with LLM verification for semantic filtering~\citep{zheng2024judging, gu2024survey}.

%% file: sections/04_formulation.tex
\section{Population-Level Alignment Protocol}
\label{sec:formulation}

We measure population-level alignment by comparing the attribute distributions of a synthetic corpus against those of a reference corpus. 
Let $X=\{x_n\}_{n=1}^{N}$ be the reference corpus, where each $x_n$ is one complete dialogue, and let $\hat{X}=\{\hat{x}_m\}_{m=1}^{M}$ be the synthetic corpus. 
A feature extractor $\phi$ maps each dialogue to $D$ dialogue-level attributes,
\[
\phi(x)=(\phi_1(x),\ldots,\phi_D(x)).
\]
We partition these attributes into behavioral attributes $B$ and structural attributes $S$. 
Behavioral attributes capture user-interaction patterns, such as primary intent, interaction mode, and politeness strategy. 
Structural attributes capture dialogue form, such as turn count and utterance length.
For each attribute $d$, we estimate its empirical distribution in the reference corpus, $p_d$, and in the synthetic corpus, $\hat{p}_d$. 
Categorical attributes are counted directly. 
Continuous structural attributes are discretized using quantile bins computed from the reference corpus and applied to both corpora.

We quantify distributional mismatch with Jensen--Shannon divergence (JS), averaged separately over behavioral and structural attributes:
\begin{align}
\mathrm{Behav\text{-}JS} &= \frac{1}{|B|}\sum_{d \in B} \mathrm{JS}(p_d, \hat{p}_d), \\
\mathrm{Struct\text{-}JS} &= \frac{1}{|S|}\sum_{d \in S} \mathrm{JS}(p_d, \hat{p}_d).
\end{align}
Lower values indicate closer population alignment. 
We use $\mathrm{Behav\text{-}JS}$ as the primary alignment metric and $\mathrm{Struct\text{-}JS}$ as a structural check, testing whether behavioral gains preserve dialogue form.

%% file: sections/05_method.tex
\section{GroupPersona: Group-Conditioned Synthetic Dialogue Generation}
\label{sec:method}

GroupPersona turns reference-corpus behavior patterns into generation controls for the user agent. 
As shown in Figure~\ref{fig:framework}, it extracts source attributes, reduces each dialogue to a core behavioral signature using verified behavior dependencies, clusters the reduced signatures into behavioral groups, enriches them into user-agent profiles, and samples groups by prevalence during synthesis.

\subsection{Stage 1: Extracting Source Attributes}
\label{sec:attribute-extraction}

For each source dialogue, GroupPersona separates attributes by their role in the pipeline. 
Behavior labels define the population patterns to be aligned and are therefore used for group discovery. 
Structural statistics describe how a dialogue is realized, such as turn count and utterance length, and are reserved for profile enrichment and evaluation.

The behavior schema contains 12 dimensions spanning intent and interaction mode, user type and goal, brevity and persistence, topic cohesion and politeness, flexibility and adaptability, and repair and recovery behavior. 
These dimensions draw on intent modeling~\citep{sarikaya2017technology}, user adaptation and dialogue evaluation~\citep{cheng2024autopal, finch2020unified, mehri2020usr}, and spoken-dialogue error handling~\citep{bohus2004error, bohus2005sorry}. 
Each dialogue is represented by one attribute--value pair per behavior dimension.

This separation prevents group discovery from collapsing into surface-form clusters. 
If structural statistics were used to form groups, clusters could be driven by length or turn count rather than by recurring user-interaction patterns. 
GroupPersona instead discovers behavior-centered groups first, then adds structural statistics so the generated corpus can preserve dialogue form without letting form define group identity.

\subsection{Stage 2: Discovering Behavioral Groups}
\label{sec:rule-mining}

Stage~2 turns behavior-labelled dialogues into behavior groups through three steps: verify behavior dependencies, reduce dialogue signatures, and cluster the reduced signatures.

\paragraph{Verified rules.}
Each dialogue is represented as a set $P_x$ of 12 behavior attribute--value pairs. 
We mine association rules over the training dialogues, where each rule predicts a target behavior pair $\ell_t=(a=v)$ from one to three antecedent pairs $A$, a range chosen to keep dependencies local and interpretable:
\[
A \Rightarrow \ell_t .
\]
Candidate rules are first filtered by support, confidence, and lift, and ranked by
\[
\mathrm{score}(r)
=
\mathrm{conf}(r)\,
\log_2 \mathrm{lift}(r)\,
\sqrt{\mathrm{support}(r)}.
\]
Each surviving rule is backward-pruned to a minimal sufficient antecedent $A^\star$: an antecedent pair is removed if deleting it reduces confidence by no more than $\delta$. 
Support, confidence, lift, and score are recomputed for the minimal rule. 
Because strong rules can still reflect templates, duplicate labels, or unsupported correlations, an LLM judge verifies each minimal rule using its statistics, matched examples, and an accept/reject rubric. 
Only LLM-accepted rules can mark a target behavior pair as derivable; rejected rules never remove pairs during signature reduction. 
Thresholds, prompts, and diagnostics are in App.~\ref{app:group_diagnostics} and App.~\ref{app:llm_judge_rubric}.

\paragraph{Reduced signatures.}
Accepted rules are applied locally within each dialogue. 
If $A^\star \Rightarrow \ell_t$ is accepted and a dialogue contains both the antecedent and target pair, then $\ell_t$ is treated as derivable in that dialogue and removed from its clustering signature. 
Rules are applied until no additional pair can be removed:
\[
R_x =
P_x \setminus 
\{\ell_t : \exists A^\star \Rightarrow \ell_t,\;
A^\star \subseteq P_x,\;
\ell_t \in P_x\}.
\]
The result $R_x$ is a variable-length reduced signature containing the behavior pairs that remain as candidate group identity. 
The reduction is dialogue-local. 
For example, if \emph{interaction mode = command-style} and \emph{user goal = tasker} predict \emph{response brevity = short}, the brevity pair is removed only from dialogues containing that antecedent pattern. 
Removed pairs are not discarded and can later appear as behavior tendencies in the group profile.

\paragraph{Groups.}
GroupPersona clusters reduced signatures with a fixed greedy procedure. 
At each step, it groups unassigned dialogues by identical reduced signature and selects the most common signature as the seed $R_s$. 
The seed is a full reduced signature, not a single behavior pair, so each cluster is anchored by a recurring combination of core behaviors. 
GroupPersona then assigns to this cluster all unassigned dialogues whose reduced signatures have Jaccard similarity at least $\tau_{\mathrm{jacc}}$ to the seed:
\[
J(R_x,R_s)=\frac{|R_x\cap R_s|}{|R_x\cup R_s|}
\ge \tau_{\mathrm{jacc}} .
\]
Assigned dialogues are removed from the pool, and the process repeats until all dialogues are assigned.

Each valid cluster is named by up to $K_{\max}$ root pairs. 
A root pair is a retained behavior pair $\ell=(a=v)$ that is both common inside the cluster and distinctive relative to the full training corpus:
\[
\Pr[\ell \mid C] \ge \tau_{\mathrm{hom}}, \qquad
\frac{\Pr[\ell \mid C]}{\Pr[\ell]} \ge \tau_{\mathrm{lift}}.
\]
Roots are selected after clustering and only from reduced signatures, so they name core group identity, not rule-derivable consequences. 
Clusters failing the size or root-admission criteria are treated as residuals and merged into the nearest valid group for synthesis. 
We use the locked configuration $\tau_{\mathrm{jacc}}=0.50$, $\tau_{\mathrm{hom}}=0.60$, $\tau_{\mathrm{lift}}=1.15$, $\tau_{\mathrm{size}}=3$, and $K_{\max}=4$. 
Details are in App.~\ref{app:group_diagnostics}.

\subsection{Stage 3: Enriching Group Profiles}
\label{sec:root-selection}

Each valid behavioral group is converted into a user-agent profile with three fields. 
\emph{Behavior roots} are the admissible root pairs from Stage~2 and define the group identity. 
\emph{Behavior tendencies} summarize frequent non-root behavior pairs within the group, including pairs removed during signature reduction. 
These tendencies do not define membership, but preserve recurring consequences and contextual signals for generation. 
\emph{Structural statistics} summarize dialogue form, such as turn count, word count, and utterance length.

Thus, the profile separates identity from realization. 
Roots specify the population behavior represented by the group, tendencies describe accompanying behaviors, and structural statistics guide the expected dialogue shape. 
An example enriched profile in the format passed to the user agent is provided in App.~\ref{app:group_profile_example}.

\subsection{Stage 4: Group-Conditioned Synthesis}
\label{sec:synthesis}

For each synthetic dialogue, GroupPersona selects a behavioral group according to its prevalence in the reference-corpus training split and retrieves a source prefix from that group. 
The user agent receives the prefix and the selected group profile. 
The assistant agent receives only the visible dialogue history and follows a fixed generic-assistant prompt.

Group selection controls corpus composition, while profile conditioning controls user behavior. 
The assistant remains group-blind and follows a fixed prompt throughout synthesis. 
Full prompts and decoding settings are provided in App.~\ref{app:prompts}.

%% file: sections/06_dataset.tex
\section{Evaluation Setup: PopAlign-Bench}
\label{sec:benchmark}

\begin{figure*}[t]
\centering
\includegraphics[width=0.96\textwidth]{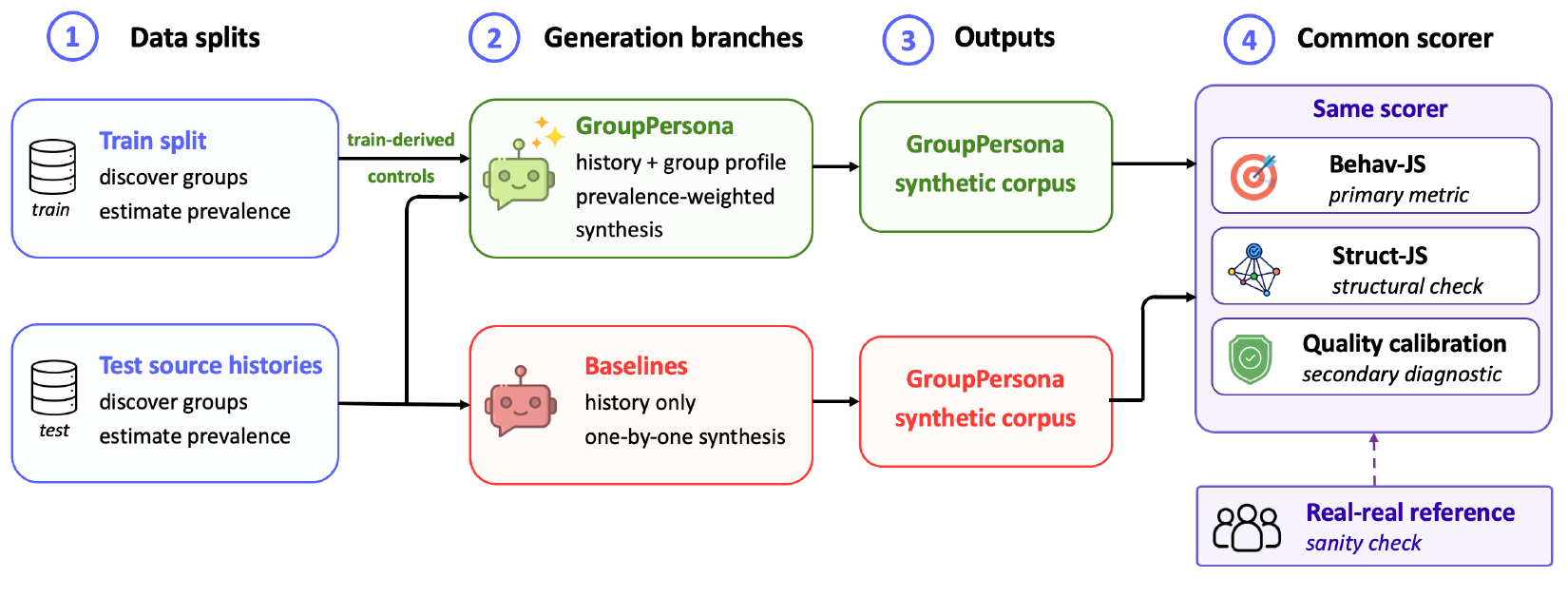}
\caption{\textbf{Evaluation protocol.} GroupPersona uses train-derived group profiles, while baselines synthesize from each test history alone. All outputs are scored with the same Behav-JS, Struct-JS, quality-calibration, and reference--reference diagnostic pipeline.}
\label{fig:evaluation_protocol}
\end{figure*}

To evaluate population-level alignment, we need corpora that are grounded in public dialogue data, share a common user--assistant format, and expose diverse user behavior distributions. 
Existing sources alone do not satisfy all three needs: task-oriented and knowledge-grounded datasets provide controlled dialogue settings but limited user-behavior variation, while Reddit discussions provide richer social behavior but are not assistant interactions. 
We therefore construct \textbf{PopAlign-Bench} from MultiWOZ~2.2~\citep{zang2020multiwoz}, Wizard of Wikipedia~\citep{dinan2018wizard}, and Pushshift Reddit~\citep{baumgartner2020pushshift}.

PopAlign-Bench crosses two source families with two construction variants. 
The \textbf{Public} family combines MultiWOZ~2.2 and Wizard of Wikipedia; the \textbf{Reddit} family uses Pushshift discussions. 
For each family, \textbf{Basic} preserves the source dialogue organization with minimal format standardization, while \textbf{Advanced} rewrites the same source conversations into a controlled user--assistant format with more varied wording, turn realization, and user-side interaction style. 
This yields four corpora: \textbf{Public-Basic}, \textbf{Public-Advanced}, \textbf{Reddit-Basic}, and \textbf{Reddit-Advanced}. 
Each corpus contains 1,000 training conversations and 500 test conversations with fixed test source histories.

\paragraph{Splits and scorer.}
The train split is used for group discovery and group-prevalence estimation. 
The test split is used for reported model comparisons. 
All methods generate conversations from the same fixed test source histories and are scored by the same labelling and extraction pipeline, as summarized in Figure~\ref{fig:evaluation_protocol}. 
The scorer reports Behav-JS and Struct-JS as defined in \S\ref{sec:formulation}. 
We also report a reference--reference value between two train-split samples as a metric sanity check: it should be small for samples from the same reference variant, but not exactly zero because of sampling variation. 
Finally, LLM-as-Judge quality calibration is reported as a secondary diagnostic, measuring whether synthetic conversations stay close to test-conversation quality scores rather than simply scoring higher.

%% file: sections/07_experiments.tex
\section{Experiments}
\label{sec:experiments}

We evaluate whether GroupPersona improves population-level alignment, whether discovered behavioral groups explain the gain, which profile fields control synthesis, and whether outputs stay calibrated to reference-conversation quality.

\paragraph{Comparison protocol.}
GroupPersona is a corpus-level method: it uses the train split to discover behavioral groups, estimate group prevalence, and build group profiles. 
At evaluation time, all methods synthesize conversations for the same fixed test source histories. 
The baselines synthesize one conversation per test history using their method-specific persona, profile, or prompting strategy, without train-derived group profiles or prevalence estimates.
This isolates the central comparison: local history/persona conditioning versus population-level control.

We compare against seven baselines spanning persona/profile-conditioned generation and prompting:
AutoPAL~\citep{cheng2024autopal},
ConceptPersona~\citep{kim2023concept},
DiaSynth~\citep{suresh2025diasynth},
FaithfulPersona~\citep{jandaghi2024faithful},
ICL~\citep{brown2020language},
Interlocutor~\citep{occhipinti2025harry},
and PersonaLens~\citep{zhao2025personalens}.
All methods run on the four PopAlign-Bench corpora and follow the evaluation protocol in Figure~\ref{fig:evaluation_protocol}. 
They are scored under the same labelling family within each cell; unless otherwise noted, main-text results use Claude Sonnet 4 for behavior labelling. 
Baseline adaptations, prompts, and decoding settings are detailed in App.~\ref{app:baselines}; bootstrap confidence intervals and seed-level details are in App.~\ref{app:bootstrap}.

\subsection{Q1. Does GroupPersona improve population-level alignment?}
\label{sec:exp-q1}

Q1 evaluates the headline claim: whether train-derived group controls make the generated corpus closer to the test population distribution. 
We report the primary Behav-JS comparison, then check structure preservation, external-label transfer, and robustness across LLM families.

\input{tables/main/main_methods_comparison_table}

\paragraph{Main result.}
Table~\ref{tab:main-eval} compares each generated corpus against the test distribution. 
GroupPersona achieves the lowest average Behav-JS, reducing it from $0.234$ for the strongest average baseline to $0.177$, a $24.4\%$ reduction. 
It is best or tied-best on all four corpora. 
The largest reductions appear on Public-Basic, where Behav-JS falls from $0.232$ to $0.148$, and Reddit-Advanced, where it falls from $0.269$ to $0.174$. 
GroupPersona also improves Public-Advanced, reducing Behav-JS from $0.252$ to $0.228$, and matches the best baseline on Reddit-Basic ($0.156$ vs.\ $0.156$). 
Together, these results show that train-derived population controls improve behavioral alignment across the full benchmark.

\paragraph{Behavioral gains preserve form and pass external checks.}
GroupPersona's average Struct-JS is $0.167$, comparable to the best baseline average, FaithfulPersona at $0.169$, indicating that the behavioral gain does not come from degrading dialogue form. 
Full per-corpus structural and external results are reported in App.~Table~\ref{tab:full-alignment}.

We further report two external checks, neither used for group construction. 
\textbf{Ext-Act} averages JS over MIDAS and DailyDialog dialogue-act labels, providing an independent view of interaction behavior, including questions, commands, statements, thanks, and apologies~\citep{yu2019midas, li2017dailydialog}. 
\textbf{Ext-E/T} averages JS over DailyDialog emotion and topic labels, serving as a drift check for dimensions outside our 12 behavior attributes.

GroupPersona reduces Ext-Act by $23\%$ over the best-average baseline, showing that its behavioral alignment generalizes to act taxonomies. 
On Ext-E/T, all methods remain close, with averages between $0.066$ and $0.085$; GroupPersona stays within this stable range, indicating that its behavior controls do not introduce emotion or topic drift.

\paragraph{The result is robust to labelling family.}
\input{tables/main/cross_llm_table}

Table~\ref{tab:cross-llm} tests whether the headline result depends on the LLM family used for synthesis or labelling. 
Across five full-benchmark settings, GroupPersona consistently outperforms ConceptPersona, with relative Behav-JS reductions of $17.6\%$--$24.7\%$. 
The gains hold when synthesis and labelling use the same family and when Claude Sonnet 4 outputs are independently relabelled by DeepSeek-R1 or Llama-3. 
In the triple-family Reddit-Advanced setting, where rewriting, synthesis, and labelling use different families, the reduction remains $27.2\%$. 
Thus, the gain is not tied to a single LLM family.

\subsection{Q2. Do the discovered behavioral groups explain the gain?}
\label{sec:exp-q2}

\input{tables/main/ablation_main}

Q2 separates useful group discovery from superficial group conditioning. 
If any group label were enough, random or $k$-means groups should perform similarly; if statistical rule mining were enough, removing the LLM verifier should not hurt. 
Table~\ref{tab:ablation-full} tests both alternatives, with full per-corpus results in App.~Table~\ref{tab:ablation-full-appendix}.

\paragraph{Discovered groups are necessary.}
Table~\ref{tab:ablation-full} shows that replacing GroupPersona's groups weakens behavioral alignment while keeping the synthesis pipeline unchanged. 
Average Behav-JS increases from $0.177$ to $0.207$ with $k$-means groups and to $0.245$ with random groups. 
Thus the gain is not from adding a generic group identifier; it comes from discovering compact behavioral signatures.

The discovered groups also have practical granularity. 
GroupPersona produces $96/104/88/73$ groups across the four corpora, with residual rates between $7.5\%$ and $10.2\%$. 
This gives enough groups to preserve population heterogeneity without leaving many dialogues in residual clusters. 
Thresholds are selected on a held-out subset of the train split; the test split is never used for selection.

\paragraph{LLM-verified reduction improves grouping.}
GroupPersona removes a behavior pair only when an LLM-accepted rule marks it as a predictable consequence. 
For example, an accepted rule may link \emph{interaction mode = command-style} and \emph{user goal profile = tasker} to \emph{response brevity = short}; then the short-brevity pair is removed from the clustering signature. 
By contrast, a high-score rule linking \emph{response brevity = short} to \emph{topic cohesion = fragmented} is rejected if it reflects a construction-specific co-occurrence rather than a behavioral dependency, so it removes nothing.

This verifier gate matters empirically: removing it worsens average Behav-JS from $0.177$ to $0.189$. 
A human audit further supports this choice: behavior labelling is stable across model families ($4.07$--$4.11$/5), and Claude Sonnet 4 gives the strongest rule-verification agreement ($3.83 \pm 0.12$; App.~\ref{app:human_audit}). 
We therefore use Claude Sonnet 4 as the default verifier and only its accepted rules remove derivable labels. 
Detailed diagnostics are in App.~\ref{app:group_diagnostics}.

\subsection{Q3. Which profile signals matter for synthesis?}
\label{sec:exp-q3}

Q3 isolates synthesis-time controls after group discovery is fixed. We remove one input at a time: source prefix, behavior roots, behavior tendencies, or structural statistics.

\paragraph{Structure statistics and behavior roots carry most control.}
Removing structural statistics causes the largest Behav-JS degradation (+0.066), followed by behavior roots (+0.041), source prefix (+0.023), and behavior tendencies (+0.009). Thus, structural statistics help realize the selected group in the target dialogue form, while roots specify the group identity.

\paragraph{The profile is most stable.}
Some ablations improve isolated cells, but none dominates across behavior, structure, and checks, making the full profile the safest cross-corpus configuration.

\subsection{Q4. Quality-score calibration to reference conversations}
\label{sec:exp-q4}

Higher LLM-as-Judge scores are not always better: an over-polished corpus can score above reference conversations while becoming less realistic as a population proxy. We therefore compute mean absolute deviation (MAD) from the reference-conversation quality profile under the same 8-dimension judge rubric; lower is better. Full rubric, per-dimension scores, and per-corpus MAD results are in App.~\ref{app:quality_calibration}.

\input{tables/main/gt_calibration_table}

\paragraph{GroupPersona best matches the reference-conversation quality range.}
Table~\ref{tab:gt-calibration} shows that reference conversations occupy a mid-range quality profile, with an average judge score of $6.93$.
Several persona-grounded baselines overshoot this range, while AutoPAL and Interlocutor fall below it.
ICL is the closest baseline with MAD $0.91$, but GroupPersona is closer with MAD $0.63$, a $31\%$ reduction.
Thus, GroupPersona improves population fidelity without merely producing judge-preferred polish.

%% file: tables/main/main_methods_comparison_table.tex
\begin{table*}[t]
\centering
\footnotesize
\setlength{\tabcolsep}{3.6pt}
\renewcommand{\arraystretch}{1.07}
\caption{Distribution alignment on PopAlign-Bench under Claude Sonnet 4 labelling. Per-corpus columns report Behav-JS; average columns add Struct-JS and external checks. Lower is better.}
\label{tab:main-eval}
\begin{tabular}{@{}lcccccccc@{}}
\toprule
& \multicolumn{4}{c}{\textbf{Behav-JS by corpus}}
& \multicolumn{4}{c}{\textbf{Average checks}} \\
\cmidrule(lr){2-5}
\cmidrule(lr){6-9}
\textbf{Method}
& Public-B & Public-A & Reddit-B & Reddit-A
& Behav & Struct & Ext-Act & Ext-E/T \\
\midrule
AutoPAL
& 0.244 & 0.256 & 0.187 & 0.274
& 0.240 & 0.170 & 0.167 & 0.073 \\

ConceptPersona
& 0.236 & \underline{0.252} & 0.181 & \underline{0.269}
& 0.235 & 0.177 & 0.154 & 0.070 \\

DiaSynth
& 0.247 & 0.259 & \underline{0.162} & 0.288
& 0.239 & 0.176 & 0.155 & 0.070 \\

FaithfulPersona
& 0.235 & 0.263 & \textbf{0.156} & 0.280
& \underline{0.234} & \underline{0.169} & \underline{0.145} & 0.072 \\

ICL
& 0.327 & 0.323 & 0.310 & 0.316
& 0.319 & 0.231 & 0.248 & 0.085 \\

Interlocutor
& \underline{0.232} & 0.254 & 0.265 & 0.284
& 0.258 & 0.183 & 0.186 & 0.081 \\

PersonaLens
& 0.245 & 0.271 & 0.210 & 0.284
& 0.253 & 0.176 & 0.168 & \textbf{0.066} \\
\midrule
\textbf{GroupPersona}
& \textbf{0.148} & \textbf{0.228} & \textbf{0.156} & \textbf{0.174}
& \textbf{0.177} & \textbf{0.167} & \textbf{0.112} & \underline{0.068} \\
\midrule
\textit{Reference}
& \textit{0.030} & \textit{0.024} & \textit{0.029} & \textit{0.023}
& \textit{0.026} & \textit{0.085} & --- & --- \\
\bottomrule
\end{tabular}
\end{table*}

%% file: tables/main/cross_llm_table.tex
\begin{table}[t]
\centering
\footnotesize
\setlength{\tabcolsep}{3pt}
\renewcommand{\arraystretch}{1.10}
\caption{
Cross-family robustness on Behav-JS ($\downarrow$).
R/S/L denotes rewriter/synthesizer/labeller: C = Claude Sonnet 4, D = DeepSeek-R1, L = Llama-3.
Reduction is relative to ConceptPersona ($\uparrow$).
}
\label{tab:cross-llm}
\begin{tabularx}{\columnwidth}{l l c c c}
\toprule
\textbf{Setting} & \textbf{R/S/L} & \textbf{GroupP.} & \textbf{ConceptP.} & \textbf{Red.} \\
\midrule
\multicolumn{5}{l}{\textit{In-family synth. + label}} \\
Claude Sonnet 4    & C/C/C & 0.177 & 0.235 & 24.7\% \\
DeepSeek-R1 & C/D/D & 0.187 & 0.238 & 21.4\% \\
Llama-3     & C/L/L & 0.197 & 0.239 & 17.6\% \\
\midrule
\multicolumn{5}{l}{\textit{Out-of-family relabel}} \\
C$\to$D & C/C/D & 0.177 & 0.234 & 24.4\% \\
C$\to$L & C/C/L & 0.184 & 0.235 & 21.7\% \\
\midrule
\multicolumn{5}{l}{\textit{Triple-family, Reddit-Advanced}} \\
L/C/D & L/C/D & 0.177 & 0.243 & 27.2\% \\
\bottomrule
\end{tabularx}
\end{table}

%% file: tables/main/ablation_main.tex
\begin{table*}[t]
\centering
\footnotesize
\setlength{\tabcolsep}{3.4pt}
\renewcommand{\arraystretch}{1.08}
\caption{
GroupPersona ablations under Claude Sonnet 4 labelling (JS divergence, $\downarrow$).
Per-corpus columns report Behav-JS.
Average columns report Behav-JS, its change from GroupPersona-full ($\Delta$; positive is worse), Struct-JS, Ext-Act, and Ext-E/T.
}
\label{tab:ablation-full}
\begin{tabular}{@{}lccccccccc@{}}
\toprule
& \multicolumn{4}{c}{\textbf{Behav-JS by corpus}}
& \multicolumn{5}{c}{\textbf{Average checks}} \\
\cmidrule(lr){2-5}
\cmidrule(lr){6-10}
\textbf{Variant}
& Public-B & Public-A & Reddit-B & Reddit-A
& Behav & $\Delta$ & Struct & Ext-Act & Ext-E/T \\
\midrule
\textbf{GroupPersona-full}
& \textbf{0.148} & \textbf{0.228} & \textbf{0.156} & \textbf{0.174}
& \textbf{0.177} & -- & \textbf{0.167} & \textbf{0.112} & 0.068 \\

\midrule
\multicolumn{10}{@{}l}{\textit{Group-discovery ablations}} \\
\quad no LLM rule verifier
& 0.160 & 0.230 & 0.174 & 0.191
& 0.189 & +0.012 & 0.167 & 0.128 & 0.066 \\

\quad $k$-means groups
& 0.182 & 0.248 & 0.194 & 0.205
& 0.207 & +0.030 & 0.167 & 0.143 & 0.074 \\

\quad random groups
& 0.233 & 0.264 & 0.235 & 0.246
& 0.245 & +0.068 & 0.167 & 0.174 & 0.079 \\

\midrule
\multicolumn{10}{@{}l}{\textit{Profile-conditioning ablations}} \\
\quad $-$ source prefix
& 0.129 & 0.265 & 0.231 & 0.176
& 0.200 & +0.023 & 0.177 & 0.146 & 0.068 \\

\quad $-$ behavior roots
& 0.223 & 0.281 & 0.190 & 0.178
& 0.218 & +0.041 & 0.194 & 0.166 & 0.072 \\

\quad $-$ behavior tendencies
& 0.149 & 0.265 & 0.174 & 0.155
& 0.186 & +0.009 & 0.154 & 0.135 & 0.069 \\

\quad $-$ structural statistics
& 0.256 & 0.276 & 0.216 & 0.224
& 0.243 & +0.066 & 0.221 & 0.173 & 0.067 \\
\bottomrule
\end{tabular}
\end{table*}

%% file: tables/main/gt_calibration_table.tex
\begin{table}[t]
\centering
\small
\setlength{\tabcolsep}{5pt}
\renewcommand{\arraystretch}{1.05}
\caption{Quality calibration to reference conversations. Avg. averages the eight judge dimensions; MAD measures deviation from the reference-conversation quality profile. Full scores are in Table~\ref{tab:gt-calibration-full}.}
\label{tab:gt-calibration}
\begin{tabular}{lcc}
\toprule
\textbf{Method} & \textbf{Avg. score} & \textbf{MAD$\downarrow$} \\
\midrule
\textit{Reference} & \textit{6.92} & \textit{0.00} \\
\midrule
AutoPAL & 5.03 & 1.89 \\
ConceptPersona & 8.31 & 1.39 \\
DiaSynth & 8.69 & 1.77 \\
FaithfulPersona & 8.07 & 1.15 \\
ICL & 6.01 & \underline{0.91} \\
Interlocutor & 5.44 & 1.49 \\
PersonaLens & 8.49 & 1.57 \\
\midrule
\textbf{GroupPersona} & \textbf{7.45} & \textbf{0.63} \\
\bottomrule
\end{tabular}
\end{table}

%% file: sections/08_limitations.tex
\section{Conclusion}
\label{sec:conclusion}

We introduced GroupPersona, a framework for addressing behavior-distribution misalignment in synthetic dialogue corpora. While persona-grounded generators can produce locally coherent conversations, they do not necessarily preserve the behavior mix of a reference population. GroupPersona turns this corpus-level target into generation-time control by distilling reference-corpus behavior patterns into prevalence-weighted groups for user-agent synthesis, while keeping structural realization separate from group identity. Across four PopAlign-Bench corpora, GroupPersona improves behavioral alignment, preserves structure, remains robust across labelling families, and better matches reference-conversation quality. These results suggest that corpus-level behavior fidelity is not only measurable, but can be improved through group-conditioned generation.

\section*{Limitations}

Our validation has five main limits.
First, root admission uses fixed thresholds $(\tau_{\mathrm{hom}}, \tau_{\mathrm{lift}})$; adaptive thresholds may reduce residuals without increasing root arity.
Second, behavior labelling and rule verification rely on frontier LLMs, and scaling to weaker open-weight labellers remains open.
Third, PopAlign-Bench covers English task-oriented, knowledge-grounded, and social-discussion corpora; other languages, domains, and modalities require new schema grounding.
Fourth, the calibration audit in \S\ref{sec:exp-q4} is a sanity check on a single judge family (Claude Sonnet 4); stratifying calibration error by labelling family, and connecting calibration to downstream training and deployment utility, is left to future work.
Fifth, we do not include degenerate aggregate controls such as FlatStats or $K{=}1$ grouping. These could further isolate the value of granularity, but they collapse population heterogeneity into a single global profile and are less natural synthesis settings than the intermediate behavioral groups studied here.
\section*{Ethical Considerations}

We use public datasets under their existing licenses and de-identify Reddit data by removing URLs and user mentions before labelling. The synthetic dialogues do not use private user data. The LLM judge is used only for offline rule verification, not for scoring downstream user-facing outputs. Because distribution alignment preserves source-corpus behavior patterns, it may also preserve source-corpus biases; releasing per-dimension marginals makes these biases inspectable rather than hidden only in generated text.

%% file: sections/09_appendix.tex
\appendix

\section{Behavioral schema: provenance and value vocabulary}
\label{app:schema}

The 12 behavioral dimensions span three coverage axes with prior-work provenance. 
\emph{Intent dynamics} (primary intent type, interaction mode, user type) follow~\citet{sarikaya2017technology}; 
\emph{interaction management} (response brevity, user goal profile, persistence level, topic cohesion, politeness strategy, interaction flexibility) follow~\citet{cheng2024autopal}, \citet{finch2020unified}, and~\citet{mehri2020usr}; 
\emph{user--system dynamics} (persona adaptability, repair behavior, error recovery style) follow~\citet{bohus2004error,bohus2005sorry}. 
Each dimension is observable at the conversation level, discrete with a 3--5-value vocabulary, and domain-agnostic. 
The full per-dimension definitions, value vocabularies, and tie-break rules used by the LLM labeller are reproduced below.

\input{tables/appendix/llm_12dimension_labeling_prompt}

\section{PopAlign-Bench corpus construction and reference floors}
\label{app:data_generation_prompts}

The benchmark follows the $2{\times}2$ design in \S\ref{sec:benchmark}: source family (Public assistant-style dialogues vs.\ Reddit social discussions) crossed with reference-distribution regime (Basic vs.\ Advanced). 
The source-family axis changes the user population; the regime axis changes the surface realisation of that population. 
All four cells share the same schema, scoring code, reference-floor estimator, and released source-history identifiers.

\paragraph{Source-family axis.}
The \textbf{Public} family combines MultiWOZ-2.2~\citep{zang2020multiwoz} task-oriented dialogues with Wizard-of-Wikipedia~\citep{dinan2018wizard} knowledge-grounded chitchat in a fixed 70/30 split. 
This mixture covers goal-driven information seeking and casual knowledge-grounded exchange. 
The \textbf{Reddit} family is built from Pushshift Reddit~\citep{baumgartner2020pushshift} threads in \emph{AskReddit}, \emph{askphilosophy}, \emph{advice}, \emph{NoStupidQuestions}, and \emph{relationship\_advice}. 
Each conversation is one top-level post followed by the three-deepest top-rated comment chain, with handles, URLs, and subreddit names stripped. 
This family covers longer, less task-scripted opinion exchange and advice-seeking. 
The train-train cross-family Behav-JS is $0.15$--$0.16$ (Tab.~\ref{tab:cross_corpus_js}), an order of magnitude above the within-corpus reference floor, so averaging across the two families does not hide family-specific failures.

\paragraph{Reference-distribution regime.}
The \textbf{Basic} regime keeps lightly normalised source dialogues and uses them directly as the reference. 
The \textbf{Advanced} regime re-renders every turn of the same source dialogue with a Claude Sonnet 4 diversity-promoting rewrite prompt. 
The prompt preserves speaker identity, intent, and information content per turn, while varying phrasing, vocabulary, politeness markers, and turn length within reasonable bounds. 
Empirically, the rewriter changes the structural distribution (Struct-JS Basic--Advanced $=0.31$) but leaves the behavioral distribution close to the within-corpus floor (Behav-JS Basic--Advanced $=0.022$). 
We therefore use Advanced variants as controlled stress tests for surface-form drift, not as claims about naturally occurring population samples. 
The dataset-transformation prompt is reproduced below.

\input{tables/appendix/data_generation_prompts}

\paragraph{Per-cell sizing, splits, and reference floors.}
Each corpus contains $1{,}000$ training conversations and $500$ test conversations with fixed source-history identifiers. 
The train split supplies persona, example, cluster, and pattern information to generators; the test split is the JS reference. 
For each cell, the \emph{reference--reference floor} is the JS between train and test marginals under the same scoring protocol used for synthesis. 
The Behav-JS floors are $0.030 / 0.024 / 0.029 / 0.023$, and the Struct-JS floors are $0.076 / 0.087 / 0.094 / 0.086$ for Public-Basic / Public-Advanced / Reddit-Basic / Reddit-Advanced, respectively.

\paragraph{Cross-corpus JS as scale anchors.}
Table~\ref{tab:cross_corpus_js} reports JS between the train splits of the four corpora. 
Behavioral JS between Basic and Advanced variants of the same source is $0.022$, close to the within-corpus floor. 
Behavioral JS between Public and Reddit corpora is $0.15$--$0.16$, which gives the natural scale of two distinct source populations. 
GroupPersona's headline Behav-JS values sit between the within-corpus floor and this cross-population ceiling. 
In absolute terms, GroupPersona closes $25\%$ of the cross-corpus-average dynamic range from the per-corpus-best baseline averaged across corpora ($0.227$) to the noise floor ($0.027$), with larger closure on Public-Basic ($42\%$) and Reddit-Advanced ($39\%$).

\input{tables/appendix/cross_corpus_js}

\section{Baseline implementation details}
\label{app:baselines}

All baselines are adapted to the same PopAlign-Bench input-output format. 
The synthesis LLM is held fixed to Claude Sonnet 4 across baselines and GroupPersona; the same per-corpus train split supplies persona, example, or profile information, and the same test split is used as the JS reference. 
Thus differences in JS reflect how each method represents and conditions on the source population rather than which backbone or evaluator is used. 
The prompts below describe the conditioning information supplied to the synthesis LLM; the full prompt templates are reported in Appendix~\ref{app:prompts}.

\paragraph{AutoPAL~\citep{cheng2024autopal}.}
Each synthesis record is conditioned on an autonomously generated long-form user persona. 
The synthesis LLM first writes a one-paragraph free-text profile covering preferences, topics of interest, and conversational habits, then uses it as the user-agent system prompt. 
We keep the two-stage design and replace only the base LLM with Claude Sonnet 4 for parity.

\paragraph{ConceptPersona~\citep{kim2023concept}.}
Each user persona is constructed by retrieving conceptually related attributes from an attribute pool. 
Here, the pool is the union of distinct values observed in the train split's 12-dimensional behavioral labels. 
This gives ConceptPersona access to the same label vocabulary as GroupPersona, but it produces per-record attribute bags rather than per-group root signatures.

\paragraph{DiaSynth~\citep{suresh2025diasynth}.}
A topic-and-subtopic taxonomy is mined from the train split by clustering free-text topic labels into 30 topic and 90 subtopic cells. 
Synthesis is conditioned on a sampled topic-subtopic pair and a one-line user persona, with the topic seed drawn proportionally to the train-set topic marginal.

\paragraph{FaithfulPersona~\citep{jandaghi2024faithful}.}
Each user persona is a faithfulness-filtered short-text profile. 
A generation pass writes a candidate persona for one source prefix, and a verifier pass accepts only personas above a fixed consistency threshold. 
Both generator and verifier use Claude Sonnet 4 for parity.

\paragraph{ICL~\citep{brown2020language}.}
Four randomly drawn train conversations are placed in the synthesis prompt as in-context examples. 
No explicit persona or structured conditioning is provided, so this serves as a lower-information example-based baseline.

\paragraph{Interlocutor~\citep{occhipinti2025harry}.}
The user agent is conditioned on a single user persona, and the assistant agent is conditioned on a separately mined assistant persona. 
The two agents then run a multi-turn dialogue without a shared group identifier. 
We use the released two-persona pairing scheme on the train split.

\paragraph{PersonaLens~\citep{zhao2025personalens}.}
Each persona is a structured five-field profile covering occupation, domain, expertise, tone, and intent, drawn from train marginals. 
Synthesis is conditioned on the structured persona plus a sampled task slot, using the released field schema and population prompts.

\paragraph{Reproducibility.}
All baselines use the same Claude Sonnet 4 endpoint with temperature $0.7$ and a 1500-token cap. 
For each corpus and method, we synthesize $500$ records from the same seed-prefix pool under the same random seed.

\section{Full Q1 distribution-alignment results}
\label{app:full_alignment}

Table~\ref{tab:full-alignment} expands Table~\ref{tab:main-eval} by reporting per-corpus Struct-JS, Ext-Act, and Ext-E/T in addition to Behav-JS. 
The main text keeps per-corpus Behav-JS and average checks in Table~\ref{tab:main-eval} to make the primary comparison readable.

\input{tables/appendix/full_alignment_table}

\section{Full Q2--Q3 ablation results}
\label{app:full_ablation}

Table~\ref{tab:ablation-full-appendix} expands Table~\ref{tab:ablation-full} by reporting per-corpus Struct-JS, Ext-Act, and Ext-E/T for each ablation. 
The variant names follow Table~\ref{tab:ablation-full}: group-discovery ablations test the discovered grouping mechanism, and profile-conditioning ablations test synthesis-time inputs.

\input{tables/appendix/full_ablation_table}

\section{Bootstrap confidence intervals}
\label{app:bootstrap}

Per-corpus bootstrap 95\% CIs on Behav-JS are computed by resampling synthesis and test records at the conversation level (200 iterations, percentile method). 
Table~\ref{tab:bootstrap_ci} reports GroupPersona-full versus the best baseline per (corpus, labelling-family) cell. 
The GroupPersona CI is strictly disjoint from the best-baseline CI on Public-Basic under all three families, Public-Advanced under Claude Sonnet 4, and Reddit-Advanced under all three families. 
The CIs overlap on Public-Advanced under DeepSeek-R1 / Llama-3 and on Reddit-Basic, consistent with the small absolute gaps in \S\ref{sec:exp-q1}.

\input{tables/appendix/bootstrap_ci}

\section{External scoring axes}
\label{app:external_axes}

The two external columns in Tables~\ref{tab:main-eval} and~\ref{tab:ablation-full} report JS over four axes drawn from prior dialogue-annotation schemes. 
\textbf{Ext-Act} averages MIDAS dialogue acts and DailyDialog acts. 
These axes partially overlap our 12 behavior attributes through intent, interaction mode, and politeness-related behavior, so we use Ext-Act as an external interaction-behavior check. 
\textbf{Ext-E/T} averages DailyDialog emotion and topic labels. 
These axes are not part of our behavior schema and are not targeted by GroupPersona, so we use Ext-E/T as a drift check rather than as the main optimization target. 
None of these four axes is used to construct GroupPersona's groups.

\paragraph{External label spaces.}
Ext-Act averages JS over two external dialogue-act label spaces. 
MIDAS includes functional and social acts such as questions, commands, opinions, statements, greetings, thanks, apologies, and closings. 
DailyDialog includes inform, questions, directives, and commissives. 
Ext-E/T averages JS over DailyDialog emotion and topic labels. 
These external labels are used only for evaluation and are not used for group construction or synthesis.

\paragraph{MIDAS dialogue acts.}
MIDAS~\citep{yu2019midas} defines 23 per-turn dialogue-act categories, including statement, opinion, complaint, command, open-ended question, yes-no question, thanks, apology, and closing. 
The unit of analysis is the user turn, so the scored distribution is the corpus-level marginal over user turns. 
MIDAS partially overlaps our schema through interaction mode, primary intent, and politeness strategy. 
GroupPersona's cross-corpus average on MIDAS is $0.132$ versus $0.159$ for the best-average baseline.

\paragraph{DailyDialog axes.}
DailyDialog~\citep{li2017dailydialog} provides three axes: a 4-way act axis, a 7-way emotion axis, and a 10-way topic axis. 
The act axis partially overlaps intent and interaction-mode dimensions, while emotion and topic are outside the conditioning schema. 
We read DailyDialog emotion and topic as non-degradation checks, since no baseline in our benchmark directly conditions on them.

\paragraph{Extraction protocol.}
For every corpus-method cell, both synthetic and test conversations are labelled under each external schema using a Claude Sonnet 4 labeller with the published rubric and value vocabulary. 
MIDAS labels are assigned per user turn; DailyDialog labels are assigned as dominant conversation-level act, emotion, and topic. 
We compute the empirical marginal over the relevant unit and then compute JS between synthetic and test marginals. 
The Ext-Act column reports the mean of MIDAS act and DailyDialog act; the Ext-E/T column reports the mean of DailyDialog emotion and DailyDialog topic.
As a calibration check, relabelling a 200-conversation subset with DeepSeek-R1 changes per-axis JS by at most $0.012$.

\section{Per-dimension behavioral JS radar}
\label{app:main_perdim}

\begin{figure*}[!htbp]
\centering
\includegraphics[width=\textwidth]{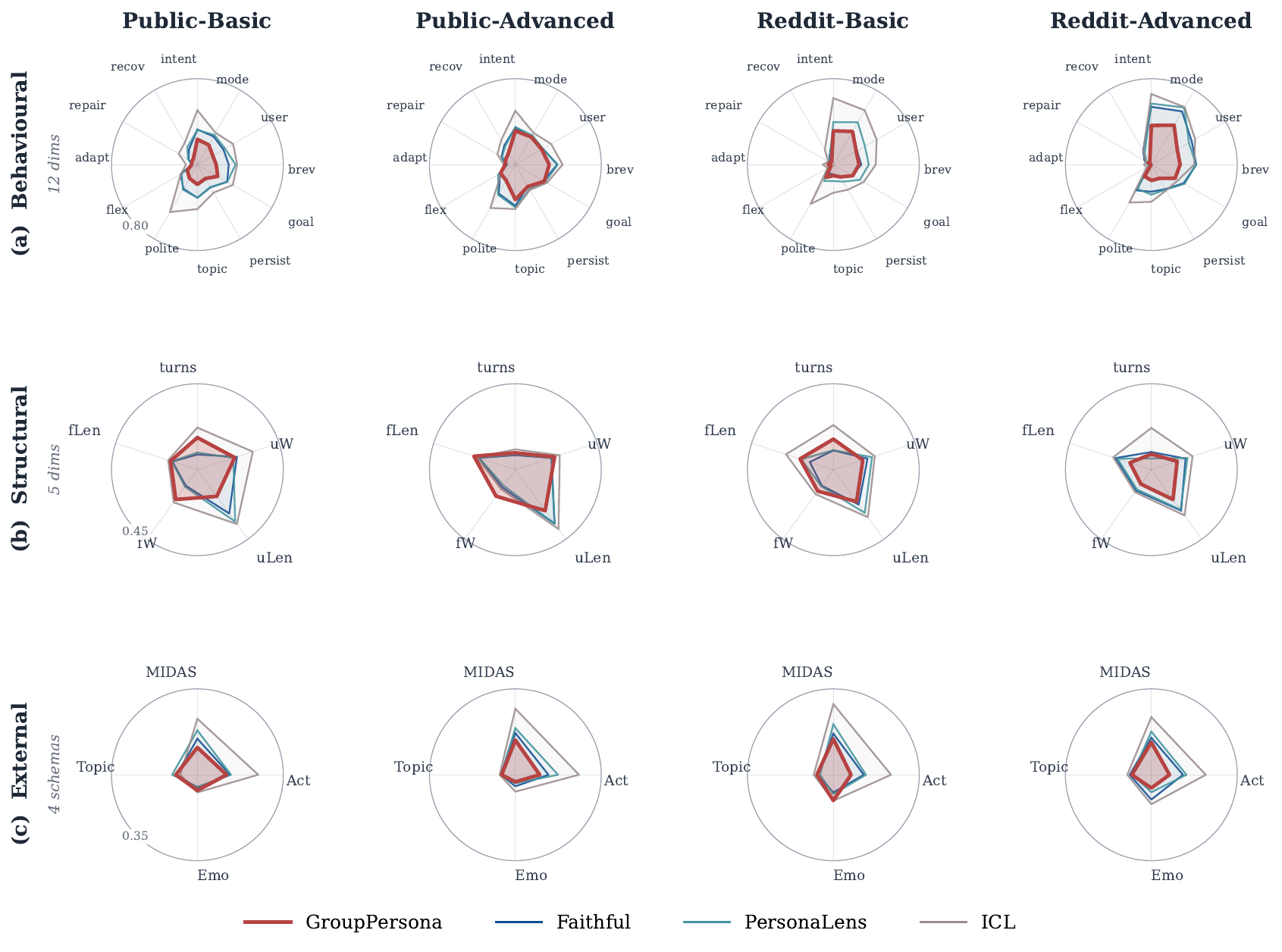}
\caption{
Per-dimension JS radar for GroupPersona-full and three representative baselines under Claude Sonnet 4. 
Rows show Behav-JS, Struct-JS, Ext-Act, and Ext-E/T; columns show the four corpora. 
GroupPersona's advantage concentrates on stylistic-adaptive behavioral dimensions and partially overlapping external act axes.
}
\label{fig:main_perdim_radar}
\end{figure*}

Figure~\ref{fig:main_perdim_radar} disaggregates the Behav, Struct, Ext-Act, and Ext-E/T columns of Table~\ref{tab:main-eval}.
The figure shows that GroupPersona's behavioral advantage concentrates on stylistic-adaptive dimensions such as \texttt{response\_brevity}, \texttt{interaction\_flexibility}, and \texttt{persona\_adaptability}. 
The identity-style dimensions are already approximately matched by strong baselines. 
On structural axes, methods sit in a tight band. 
On external axes, GroupPersona improves the partially overlapping MIDAS and DailyDialog act axes, while DailyDialog emotion and topic remain close across methods.

\section{Cross-family configurations}
\label{app:cross_family}

Table~\ref{tab:cross_family_full} expands the cross-family comparison in Table~\ref{tab:cross-llm}. 
We evaluate five configurations: three in-family settings, where the same model family performs synthesis and labelling, and two out-of-family settings, where Claude Sonnet 4 synthesizes the records and another family labels them. 
The in-family settings use Claude Sonnet 4, DeepSeek-R1, and Llama-3. 
The out-of-family settings use Claude Sonnet 4$\to$DeepSeek-R1 and Claude Sonnet 4$\to$Llama-3.

GroupPersona remains ahead of ConceptPersona on all four corpora in every configuration except Llama-3 in-family on Public-Advanced, where the gap falls within the cross-family variation observed across labellers. 
The cross-family results therefore support the main conclusion that the alignment gain is not an artefact of one labelling family.

\input{tables/appendix/cross_family_pipeline}

\section{Per-dimension breakdown of cross-family results}
\label{app:cross_model_perdim}

\begin{figure}[!htbp]
\centering
\includegraphics[width=\columnwidth]{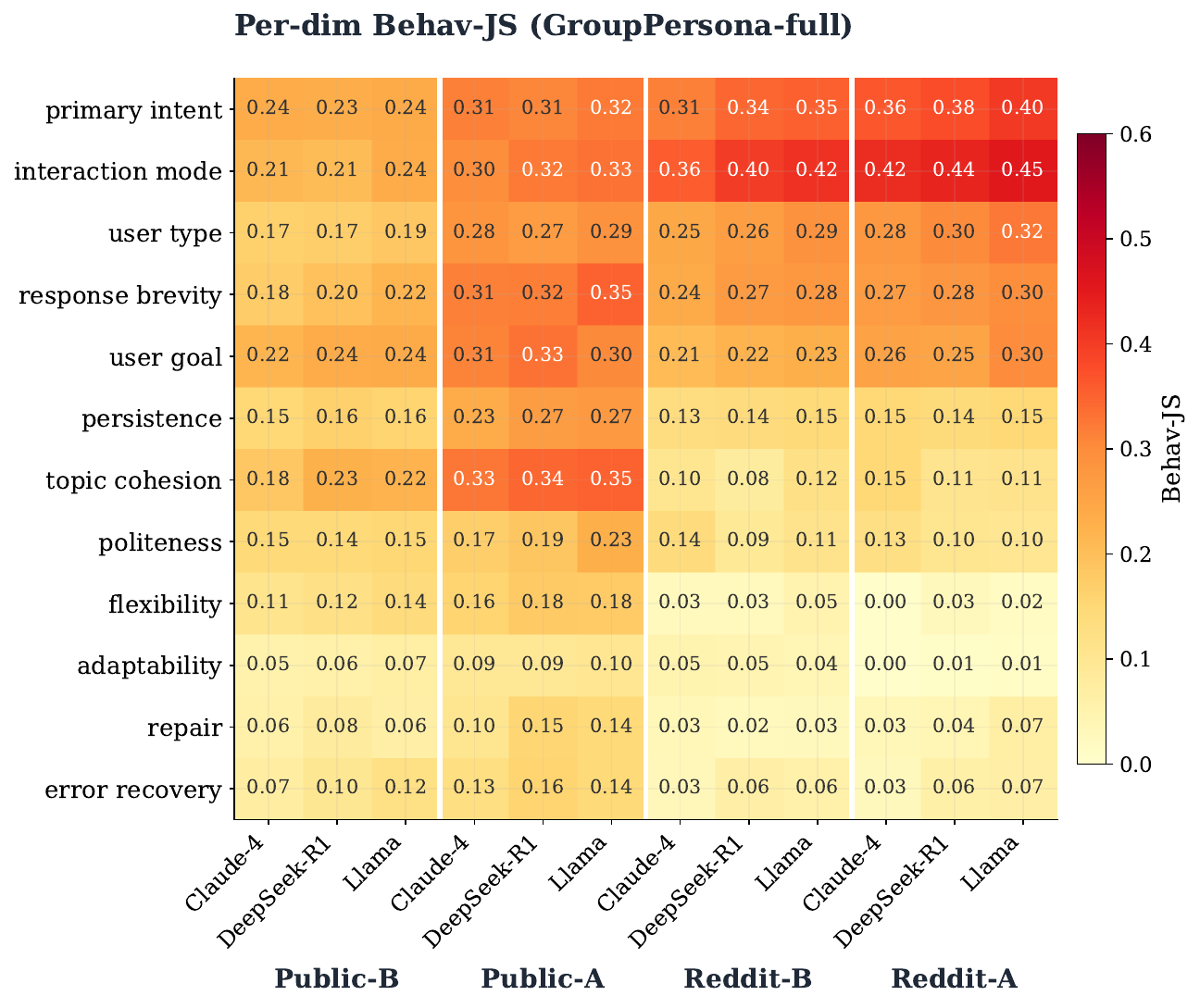}
\caption{
Per-dimension cross-family Behav-JS for GroupPersona and ConceptPersona. 
Differences across LLM families are distributed across dimensions, while GroupPersona remains consistently lower on average.
}
\label{fig:cross_model_perdim}
\end{figure}

Figure~\ref{fig:cross_model_perdim} provides the per-corpus and per-family breakdown behind the cross-family results in Table~\ref{tab:cross-llm}. 
The Claude Sonnet 4 / DeepSeek-R1 / Llama-3 gap is distributed across dimensions rather than concentrated in a single label.

\section{Group-discovery diagnostics and locked configuration}
\label{app:group_diagnostics}

This appendix reports the implementation details behind Stage~2, including rule thresholds, minimal-rule pruning, LLM verification, reduced-signature construction, greedy Jaccard clustering, root admission, residual handling, and diagnostics.

\paragraph{Verified rules.}
Each dialogue starts as a set $P_x$ of 12 behavior attribute--value pairs. 
We mine association rules over the training split with antecedent arity 1--3. 
A rule predicts a target behavior pair $\ell_t=(a=v)$ from antecedent pairs $A$, written as $A \Rightarrow \ell_t$. 
Candidate rules must satisfy confidence $\ge 0.80$, lift $\ge 1.3$, and support $\ge 0.03$. 
They are scored by
\[
\mathrm{score}(r)
=
\mathrm{conf}(r)
\log_2 \mathrm{lift}(r)
\sqrt{\mathrm{support}(r)}.
\]
For each multi-pair antecedent, we backward-prune to a minimal sufficient subset $A^\star$. 
An antecedent atom is removed when its deletion changes confidence by at most $\delta=0.05$. 
Rules that collapse to the same $(A^\star,\ell_t)$ are deduplicated. 
Each deduplicated minimal rule is then judged by the LLM verifier described in App.~\ref{app:llm_judge_rubric}. 
Only accepted rules can mark a target pair as derivable.

\paragraph{Reduced signatures.}
Accepted rules are applied independently within each dialogue. 
Rules are scanned in increasing antecedent arity and applied to a fixed point. 
For a dialogue $x$, if an accepted rule $A^\star \Rightarrow \ell_t$ satisfies $A^\star \subseteq P_x$ and $\ell_t \in P_x$, then $\ell_t$ is removed from the clustering signature. 
The resulting reduced signature is
\[
R_x =
P_x \setminus 
\{\ell_t :
\exists A^\star \Rightarrow \ell_t,
A^\star \subseteq P_x,
\ell_t \in P_x
\}.
\]
The size of $R_x$ can vary across dialogues because different rules fire in different local contexts.

\paragraph{Human audit of LLM-driven steps.}

GroupPersona uses LLMs in two places: assigning per-dialogue behavior labels and verifying whether statistically strong association rules reflect meaningful behavioral dependencies. 
We audit both steps to check whether the labels are stable across model families and whether the rule verifier provides reliable accept/reject decisions.

\input{tables/appendix/judge_role_table}

The audit shows stable per-dialogue behavior labelling across Claude Sonnet 4, DeepSeek-R1, and Llama-3, with mean label-quality scores between $4.07$ and $4.11$ on a 1--5 scale. 
For rule verification, Claude Sonnet 4 obtains the highest agreement score ($3.83 \pm 0.12$), so we use Claude Sonnet 4 as the default verifier in the main experiments.

\paragraph{Greedy Jaccard clustering.}
Dialogues are clustered in the reduced-signature space using a deterministic single-pass procedure. 
Among unassigned dialogues, we select the most frequent remaining reduced signature as the seed $R_s$. 
The cluster absorbs every unassigned dialogue $x$ whose reduced signature satisfies
\[
J(R_x,R_s)
=
\frac{|R_x \cap R_s|}{|R_x \cup R_s|}
\ge
\tau_{\mathrm{jacc}}.
\]
Absorbed dialogues are removed from the pool, and the process repeats until all dialogues are assigned. 
Ties are resolved using a fixed ordering.

\paragraph{Root admission and residual handling.}
After clustering, each valid cluster is named by up to $K_{\max}$ root pairs. 
A root pair is a retained behavior pair $\ell=(a=v)$ that is both common inside cluster $C$ and distinctive relative to the full training corpus:
\[
\Pr[\ell \mid C] \ge \tau_{\mathrm{hom}},
\qquad
\frac{\Pr[\ell \mid C]}{\Pr[\ell]} \ge \tau_{\mathrm{lift}}.
\]
Clusters smaller than $\tau_{\mathrm{size}}$ or without any admissible root pair are placed in the residual set. 
At synthesis time, each residual cluster is merged into the nearest valid group by average Jaccard similarity.

\paragraph{Locked configuration and diagnostics.}
The locked configuration is
\[
(\tau_{\mathrm{hom}}, \tau_{\mathrm{lift}}, \tau_{\mathrm{jacc}}, \tau_{\mathrm{size}}, K_{\max})
=
(0.60, 1.15, 0.50, 3, 4).
\]
For threshold selection, each corpus holds out $200$ of the $1{,}000$ train conversations for validation. 
The remaining $800$ train conversations are used for rule mining, LLM verification, signature reduction, and clustering; the validation subset is used only to score validation Behav-JS. 
The $500$-conversation test split is never used for configuration selection.

Table~\ref{tab:group-diagnostics} summarizes the locked configuration and validation sweep. 
\textbf{Rules} is the number of deduplicated minimal rules submitted to the LLM judge, \textbf{Resid.} is the residual-cluster rate, and \textbf{Drop} is the average number of behavior pairs removed per dialogue during signature reduction. 
Claude Sonnet 4 accept rates on the submitted minimal rules are $65.1\%$, $67.4\%$, $60.5\%$, and $53.5\%$ for Public-Basic, Public-Advanced, Reddit-Basic, and Reddit-Advanced, respectively.

\input{tables/appendix/group_diag_table}

\begin{figure*}[!htbp]
\centering
\includegraphics[width=\textwidth]{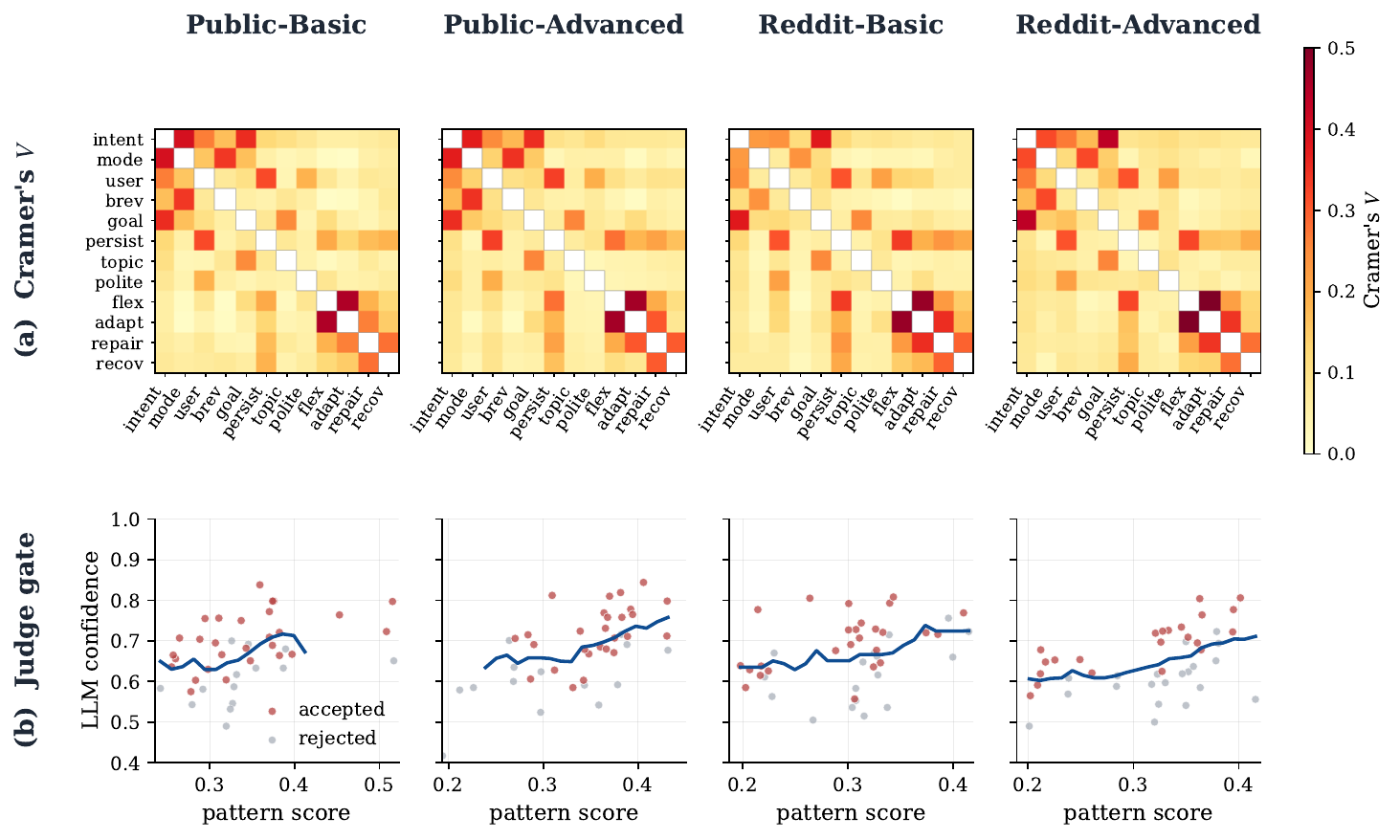}
\caption{
Group-discovery diagnostics by corpus. 
Top row: Cramér's $V$ over the 12 behavioral dimensions, showing the dependency structure exploited by local signature reduction. 
Bottom row: LLM-judge confidence versus pattern score for minimal rules. 
The judge is broadly monotone in score but filters a non-trivial set of artefactual or tautological rules, controlling which rules can remove behavior pairs during reduction.
}
\label{fig:group_diagnostics}
\end{figure*}

Figure~\ref{fig:group_diagnostics} visualizes the two diagnostics behind Stage~2: dependency structure among behavior dimensions and the LLM judge's filtering behavior over mined minimal rules.

\subsection{Human audit of LLM-driven steps}
\label{app:human_audit}

We conducted an internal human audit of the two LLM-driven steps used in group construction: per-dialogue behavior labelling and minimal-rule verification. 
The audit was conducted by project-internal expert annotators for research validation; no external participants or crowdworkers were recruited or paid, and no new personal data were collected.

Annotators reviewed complete dialogues with model-assigned behavior labels and rated whether the labels were supported by observable conversational behavior, without inferring demographics, private attributes, or unstated intent. 
They also reviewed minimal rules with their statistics and matched examples, rating whether each accept/reject decision reflected a meaningful behavior dependency rather than a template artifact, duplicate label, or unsupported correlation. 
Both audits used a 1--5 scale, where higher scores indicate stronger support or agreement.

The audit supports the LLM-driven design: behavior labelling is stable across model families ($4.07$--$4.11$/5), and Claude Sonnet 4 gives the strongest rule-verification agreement ($3.83 \pm 0.12$), so we use it as the default verifier.

\section{LLM judge rubric for minimal-rule verification}
\label{app:llm_judge_rubric}

This appendix details the LLM verification step used in Stage~2. 
After backward pruning, each candidate minimal rule is written as $A^\star \Rightarrow \ell_t$, where $A^\star$ is a minimal set of antecedent behavior pairs and $\ell_t=(a=v)$ is the target behavior pair. 
The verifier is used only to filter mined rules before signature reduction. 
It does not generate rules, cluster dialogues, or score synthetic conversations.

For each minimal rule, the judge receives four inputs: the rule in plain text, its support, confidence, lift, and pattern score, three matched training examples whose behavior-pair sets contain $A^\star$, and the parent rule before pruning. 
The matched examples mark whether $\ell_t$ is also present, allowing the judge to check whether the rule reflects a meaningful behavior dependency rather than a template artifact, near-duplicate label, or unsupported correlation. 
The parent rule is included to verify that pruning did not remove a semantically necessary antecedent.

The judge returns \texttt{is\_reasonable} $\in \{0,1\}$, \texttt{confidence} $\in [0,1]$, and a brief justification. 
Only rules with \texttt{is\_reasonable}=1 can mark a target pair as derivable during signature reduction in \S\ref{sec:rule-mining}. 
The full prompt, accept/reject rubric, confidence calibration table, and output format are reproduced below.

\input{tables/appendix/llm_pattern_judge_prompt}

\section{Human validation protocol}
\label{app:human_validation}

Two LLM-driven steps are validated against human ratings: per-record behavior labelling and per-rule accept/reject judgment.

\paragraph{Labeller-role audit.}
For every (corpus, source model) cell, we sample 40 labelled records and 40 minimal rules, giving $4 \times 3 \times 40 = 480$ records and $480$ minimal rules.
Records are rated for \emph{label\_quality}: whether the 12 labels assigned to the conversation are correct.
Minimal rules are rated for \emph{rule\_quality}: whether the minimal antecedent still captures a plausible user-behavior regularity for the consequent and, where applicable, whether the parent rule's pruning kept the semantics.

\paragraph{Inter-annotator agreement.}
Krippendorff's $\alpha$ over the three-annotator panel is $0.62$ for labeller-role ratings and $0.55$ for judge-role ratings. 
The judge-role agreement is lower because rule judgment is more open-ended than per-record label checking.

\section{Inter-judge agreement across labelling families}
\label{app:llm_judge_audit}

For every corpus and label-source cell, we run all three LLM judges over the minimal-rule set: Claude Sonnet 4, DeepSeek-R1, and Llama-3.
Per-minimal-rule binary agreement (Cohen's $\kappa$) against the union vote across judges is $0.66 / 0.62 / 0.59$ for Claude Sonnet 4 / DeepSeek-R1 / Llama-3.
Pairwise $\kappa$ is $0.61$ for Claude Sonnet 4--DeepSeek-R1, $0.55$ for Claude Sonnet 4--Llama-3, and $0.52$ for DeepSeek-R1--Llama-3.
Disagreement concentrates on borderline-score minimal rules: above pattern score $0.7$, all judges agree to accept in $91\%$ of cells; below $0.2$, they agree to reject in $96\%$ of cells.

\section{Admitted-group case study}
\label{app:residual_case_study}

This appendix illustrates the granularity of admitted groups produced by the Stage~2 group-discovery pipeline.
Table~\ref{tab:residual_case_study} shows two representative groups under the locked configuration: a large casual-chitchat group with the maximum $K_{\max}=4$ root pairs, and the smallest admitted group with $2$ root pairs at the size floor $\tau_{\mathrm{size}}=3$.
Residual clusters that fail size or root-admission criteria are not used as standalone synthesis groups; they are merged into the nearest valid group at synthesis time, as described in App.~\ref{app:group_diagnostics}.

\input{tables/appendix/residual_case_study_table}

\section{Example group profile}
\label{app:group_profile_example}

The following block shows an example enriched profile in the format passed to the user agent.
It separates group identity, accompanying behavior tendencies, and structural guidance.

\begin{tcolorbox}[
  colback=gray!4,
  colframe=gray!35,
  boxrule=0.4pt,
  left=4pt,
  right=4pt,
  top=4pt,
  bottom=4pt,
  title={Example enriched group profile}
]
\small

\noindent\textbf{Behavior roots.}
\begin{itemize}[leftmargin=*, itemsep=0pt, topsep=1pt]
    \item \texttt{interaction\_mode = QA-style}
    \item \texttt{user\_goal\_profile = Explorer}
    \item \texttt{topic\_cohesion = High}
\end{itemize}

\noindent\textbf{Behavior tendencies.}
\begin{itemize}[leftmargin=*, itemsep=0pt, topsep=1pt]
    \item Users usually ask follow-up questions rather than issuing one-shot commands.
    \item User turns are usually medium length and remain on the same topic.
    \item When the assistant response is incomplete, users tend to clarify the request instead of abandoning the task.
    \item Politeness markers such as ``please'' or ``thanks'' appear in a minority but recurring subset of dialogues.
\end{itemize}

\noindent\textbf{Structural guidance.}
\begin{itemize}[leftmargin=*, itemsep=0pt, topsep=1pt]
    \item Expected dialogue length: medium, typically 6--10 turns.
    \item Expected user utterance length: medium, typically one or two sentences.
    \item Expected interaction shape: initial information request followed by clarification or refinement turns.
\end{itemize}

\noindent\textbf{Source prefix.}
\begin{quote}
\small
\texttt{user: I am trying to decide which train route is better for a weekend trip.}\\
\texttt{assistant: Sure. What cities are you travelling between, and do you care more about price or travel time?}
\end{quote}

\noindent\textbf{Instruction to user agent.}
Generate the next user turns as a member of this group. Preserve the source prefix, follow the behavior roots as the main identity, use the tendencies as likely but not mandatory behavior, and keep the dialogue within the structural guidance.

\end{tcolorbox}

\section{Metadata enrichment prompt}
\label{app:metadata_prompt}

The metadata-enrichment step converts each discovered behavioral group into the three profile fields shown in Figure~\ref{fig:framework}: behavior roots, behavior tendencies, and structural statistics.
In the implementation prompt, these fields are serialized as \texttt{group\_persona}, \texttt{conversation\_patterns}, and \texttt{group\_statistics}, respectively.
The naming differs only at the prompt-template level: \texttt{group\_persona} is the natural-language rendering of the group's behavior roots, \texttt{conversation\_patterns} summarizes behavior tendencies, and \texttt{group\_statistics} stores structural statistics.

The same enrichment prompt is run on the GroupPersona, $k$-means, and random clusterings evaluated in \S\ref{sec:exp-q2}. 
Thus differences in synthesis fidelity reflect the quality of the discovered groups rather than differences in how metadata is generated.

\input{tables/appendix/metadata_enrichment_prompt}

\begin{figure*}[!htbp]
\centering
\includegraphics[width=\textwidth]{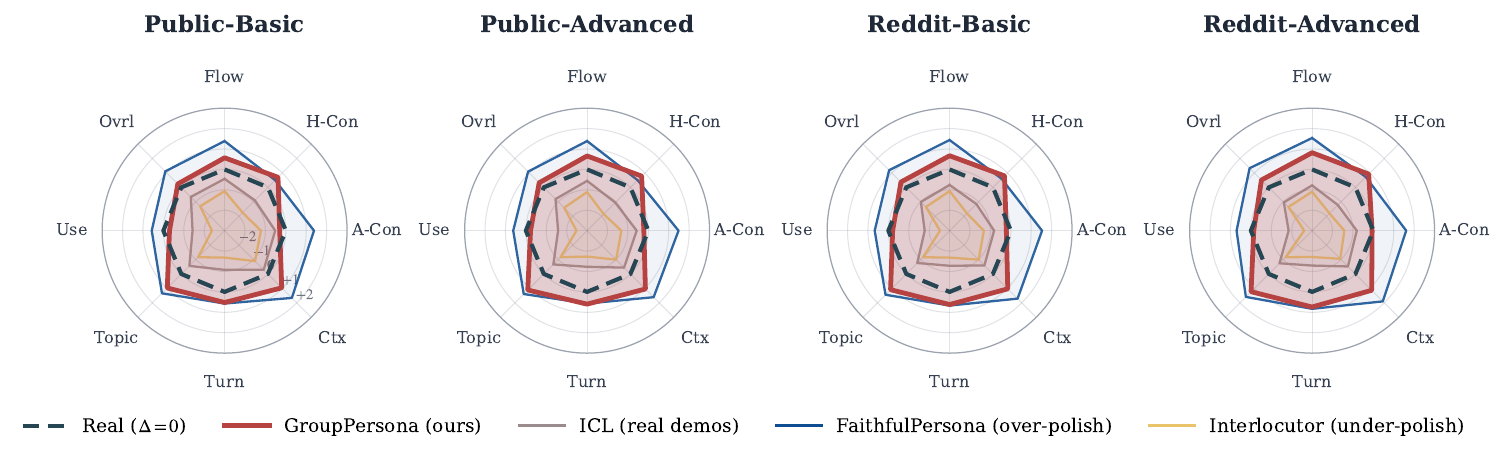}
\caption{
Per-corpus quality calibration as signed deviation from reference conversations
($\Delta=\text{Method}-\text{Reference}$) across 8 LLM-as-Judge dimensions.
The dashed circle marks the reference-conversation reference profile.
GroupPersona stays closest to the reference profile, while representative baselines show over-polishing or under-polishing.
}
\label{fig:gt_calibration_radar}
\end{figure*}

\input{tables/appendix/gt_calibration_full_table}
\section{Synthesis prompt}
\label{app:prompts}

The user-agent prompt is assembled from a fixed group index and four synthesis inputs. 
The group index identifies which group was selected, while the profile inputs correspond to the Figure~\ref{fig:framework} fields used to condition generation:
\textbf{source prefix}, \textbf{behavior roots}, \textbf{behavior tendencies}, and \textbf{structural statistics}.
In the prompt template, these are serialized as \texttt{[history]}, \texttt{[persona]}, \texttt{[patterns]}, and \texttt{[stats]}, respectively. 
Here, \texttt{[persona]} is the natural-language behavior-root guidance derived from the group's root signature; \texttt{[patterns]} contains non-root behavior tendencies; and \texttt{[stats]} contains structural statistics.

The conditioning-component ablations in \S\ref{sec:exp-q3} remove one synthesis input at a time:
\textsc{no-history} removes \texttt{[history]} and corresponds to $-$ source prefix;
\textsc{no-persona} removes \texttt{[persona]} and corresponds to $-$ behavior roots;
\textsc{no-patterns} removes \texttt{[patterns]} and corresponds to $-$ behavior tendencies;
\textsc{no-stats} removes \texttt{[stats]} and corresponds to $-$ structural statistics.
The minimal \texttt{[root]} group identifier is retained in all profile-conditioning ablations so that group selection remains fixed; it serves only as the selected group index, not as a descriptive profile field. 
Cluster-quality ablations in \S\ref{sec:exp-q2} replace this selected group with a random or $k$-means group and rerun metadata enrichment on the new group's records.

The assistant-agent system prompt is held fixed across all methods and ablations. 
The full prompt and ablation-to-block mapping are reproduced below.

\input{tables/appendix/conversation_synthesis_prompt}

\section{Conversation-quality calibration setup}
\label{app:quality_calibration}

\input{tables/appendix/quality_calibration_prompt}
\input{tables/appendix/gt_calibration_per_corpus}

This appendix documents the LLM-as-Judge rubric, prompt, and aggregation protocol used for the quality-calibration audit in \S\ref{sec:exp-q4}.

\paragraph{Why anchor against reference conversations.}
Standard LLM-as-Judge quality rubrics can reward synthetic dialogues that are overly polished relative to reference conversations. 
In our audit, reference test conversations occupy a mid-range quality band rather than the top of the 1--10 scale. 
We therefore evaluate calibration by measuring how close each method's quality-score profile is to the reference-conversation profile under the same rubric and judge.

\paragraph{Dimensions.}
We use 8 conversation-level dimensions: Flow, Human Consistency (H-Con), Assistant Consistency (A-Con), Context utilisation (Ctx), Turn balance (Turn), Topic coherence (Topic), Usefulness (Use), and Overall conversation quality (Ovrl). 
Each dimension is rated on a 1--10 scale, where higher indicates a closer match to realistic dialogue from the same corpus, not absolute polish.

\paragraph{Prompt design.}
The judge receives the candidate conversation, a short corpus description, and three reference conversations from the same corpus as anchors. 
The prompt explicitly notes that reference conversations can include hesitation, topic shifts, false starts, repair, persistence, and varied turn lengths. 
This prevents the judge from treating smoothness alone as realism. 
The full prompt is reproduced below.

\paragraph{Calibration regimes.}
The baselines fall into three broad regimes. 
ConceptPersona, DiaSynth, FaithfulPersona, and PersonaLens tend to over-polish, scoring above the reference-conversation band on many dimensions. 
AutoPAL and Interlocutor tend to fall below the band. 
ICL is the closest baseline, likely because it conditions on raw reference demonstrations. 
GroupPersona has the lowest average MAD, suggesting that population-derived behavior profiles better preserve the quality range of reference conversations.

\paragraph{Full per-dimension quality scores.}
\label{app:quality_calibration_scores}
Table~\ref{tab:gt-calibration-full} reports the full 8-dimension LLM-as-Judge scores summarized in Table~\ref{tab:gt-calibration}.
MAD is computed against the reference-conversation quality profile across the eight dimensions.

\paragraph{Per-corpus scoring and aggregation.}
For each (method, corpus) cell, we score generated conversations with the same Claude Sonnet 4 judge and average scores along each quality dimension. 
The reference-conversation row uses the same rubric applied to test conversations from the same corpus. 
Table~\ref{tab:gt-calibration} reports the average 8-dimensional score profile across corpora, while Table~\ref{tab:gt-calibration-per-corpus} reports the per-corpus MAD summary. 
Figure~\ref{fig:gt_calibration_radar} visualizes signed deviation from reference conversations for representative methods.

%% file: tables/appendix/llm_12dimension_labeling_prompt.tex
\noindent\begin{tcolorbox}[breakable, colback=gray!5, colframe=gray!40, arc=2pt,
  left=4pt, right=4pt, top=4pt, bottom=4pt, fontupper=\small,
  title={\small\textbf{12-dimension per-record labelling prompt (full)}},
  coltitle=black, colbacktitle=gray!12,
  width=\linewidth]
\footnotesize

\textbf{Role.} You are an expert in conversation behaviour analysis. Given one complete user--assistant dialogue, assign exactly one categorical label per dimension. The labels are aggregated across a large corpus for distributional analysis, so consistency of interpretation matters more than per-dialogue expressiveness.

\smallskip
\textbf{Input.} A single dialogue with turns alternating between \texttt{user:} and \texttt{assistant:}.

\smallskip
\textbf{Critical constraints.}
(1) Output \emph{only} a single valid JSON object; no Markdown, no prose, no trailing commas.
(2) Use exactly the twelve keys listed below; no missing or extra keys.
(3) Each value must be one of the allowed options for that key.
(4) If a label cannot be reliably inferred from the conversation alone, output \texttt{"unknown"}.
(5) Rely on observable behaviour; do not hallucinate user intent or demographics.

\smallskip
\textbf{Dimension definitions and value vocabularies.}

\smallskip
\textbf{1. \texttt{primary\_intent\_type}} $\in$ \{Task, Info-seeking, Chitchat, Music, Reminder, Alarm, Smart\_Home\}.
The single intent that drives the majority of the user's turns; tie-break by the opening request.

\textbf{2. \texttt{interaction\_mode}} $\in$ \{Command-style, QA-style, Multi-turn\_Dialog-style\}.
Command = imperative one-shot; QA = question $\to$ answer pattern; Multi-turn = $\geq 3$ back-and-forth turns building on prior context.

\textbf{3. \texttt{user\_type}} $\in$ \{Power\_User, One-shot\_User, Tasker, Casual\}.
Power = exploits advanced options; One-shot = ends after one exchange; Tasker = follows a task plan; Casual = informal, low engagement.

\textbf{4. \texttt{response\_brevity}} $\in$ \{Short, Medium, Long\}.
Median user-utterance word count: Short $\leq 6$, Medium $7$--$20$, Long $\geq 21$.

\textbf{5. \texttt{user\_goal\_profile}} $\in$ \{Tasker, Explorer, Goal-switcher, Chatter\}.
Tasker = single goal pursued to completion; Explorer = probing/learning; Goal-switcher = $\geq 2$ goal changes; Chatter = social, no goal.

\textbf{6. \texttt{persistence\_level}} $\in$ \{High, Medium, Low\}.
After a failed/unhelpful system reply: High = retries $\geq 3 \times$; Medium = 1--2 retries; Low = abandons.

\textbf{7. \texttt{topic\_cohesion}} $\in$ \{High, Medium, Fragmented\}.
High = one topic; Medium = related topics; Fragmented = $\geq 2$ unrelated topics.

\textbf{8. \texttt{politeness\_strategy}} $\in$ \{Direct, Polite, Formal, Friendly\}.
Direct = imperative, no marker; Polite = ``please/thanks''; Formal = third-person/honorifics; Friendly = casual greetings, emotive language.

\textbf{9. \texttt{interaction\_flexibility}} $\in$ \{Adaptive, Static\}.
Adaptive = reformulates after misunderstanding; Static = repeats verbatim or gives up.

\textbf{10. \texttt{persona\_adaptability}} $\in$ \{High, Low\}.
High = tone/style shifts to match system; Low = constant tone regardless.

\textbf{11. \texttt{repair\_behavior}} $\in$ \{Rephrase, Clarify, Ignore, Retry\}.
After a breakdown: Rephrase = restates differently; Clarify = adds info / asks back; Ignore = moves on; Retry = repeats identically.

\textbf{12. \texttt{error\_recovery\_style}} $\in$ \{Clarification\_Request, Repetition, Restart, Task\_Abandonment\}.
Dominant strategy when the system errs: ask for clarification, repeat, restart conversation, or abandon the task.

\smallskip
\textbf{Output format.}
\begin{verbatim}
{
  "primary_intent_type": "...",
  "interaction_mode": "...",
  "user_type": "...",
  "response_brevity": "...",
  "user_goal_profile": "...",
  "persistence_level": "...",
  "topic_cohesion": "...",
  "politeness_strategy": "...",
  "interaction_flexibility": "...",
  "persona_adaptability": "...",
  "repair_behavior": "...",
  "error_recovery_style": "..."
}
\end{verbatim}

\smallskip
\textbf{Tie-breaking.} If two values are equally supported, pick the one earlier in the value vocabulary. If the dialogue is too short ($<$ 4 turns) to infer a dimension, return \texttt{"unknown"} for that key rather than guess.
\end{tcolorbox}

%% file: tables/appendix/data_generation_prompts.tex
\noindent\begin{tcolorbox}[breakable, colback=gray!5, colframe=gray!40, arc=2pt,
  left=4pt, right=4pt, top=4pt, bottom=4pt, fontupper=\small,
  title={\small\textbf{Dataset-to-TOD transformation prompt (full)}},
  coltitle=black, colbacktitle=gray!12,
  width=\linewidth]
\footnotesize

This prompt converts a source dialogue (MultiWOZ / Wizard-of-Wikipedia / Pushshift Reddit thread) into a task-oriented user--assistant dialogue suitable for behavioral labelling. It runs in two sequential steps. Step~1 extracts the semantic core; Step~2 regenerates the dialogue under a target behavioural profile (one profile for the \emph{Basic} variants, four profiles for the \emph{Advanced} variants).

\smallskip
\textbf{Step 1 --- Semantic core extraction.}

\emph{Role.} You are an expert conversation analyst. Given a raw dialogue (or a Reddit thread that has already been pre-processed to remove URLs, user mentions, and subreddit references), extract the semantic core that any new conversation on the same topic must preserve.

\emph{Output (strict JSON).}
\begin{verbatim}
{
  "core_user_need": "...",
  "key_information": ["..."],
  "task_sequence": ["..."],
  "domain": "...",
  "critical_details": ["..."]
}
\end{verbatim}
\begin{itemize}
\setlength{\itemsep}{1pt}
\item \texttt{core\_user\_need}: the fundamental goal the user is trying to accomplish.
\item \texttt{key\_information}: essential facts, named entities, and context.
\item \texttt{task\_sequence}: ordered list of main sub-objectives.
\item \texttt{domain}: subject area (e.g.\ travel, cooking, technical support).
\item \texttt{critical\_details}: specific constraints, preferences, or numeric values.
\end{itemize}

\smallskip
\textbf{Step 2 --- Behavioural profile instantiation.}

\emph{Role.} You are a conversation generator. Given the semantic core from Step~1 and a target behavioural profile, produce a new user--assistant dialogue that faithfully instantiates the profile.

\emph{Constraints (apply to every variant).}
(1) Generate 10--40 alternating turns, labelled \texttt{user:} and \texttt{assistant:}.
(2) Preserve every fact in the semantic core.
(3) Do not copy more than 5 contiguous words from the source dialogue.
(4) Each turn is natural-language, no system tokens.

\emph{Profiles.}

\textbf{Basic --- short-request baseline.}
User turns are concise (1--2 sentences), task-focused, direct, no courtesy markers. Assistant responds context-appropriately.

\textbf{Advanced (four profiles, one dialogue per profile per semantic core).}
\begin{itemize}
\setlength{\itemsep}{1pt}
\item \emph{(a) Short-request, unhelpful user.} User repeatedly expresses dissatisfaction and resists assistant suggestions before grudgingly accepting one.
\item \emph{(b) Short-request, topic shift.} User makes at least two abrupt unrelated topic changes mid-conversation, returning briefly to the main task only at the end.
\item \emph{(c) Short-request, early termination.} User abandons the task before completion with an explicit closing signal (e.g.\ ``nevermind, forget it'').
\item \emph{(d) Short-request, baseline.} Standard short-request behaviour as in Basic.
\end{itemize}

\smallskip
\textbf{Reddit specifics.} For Reddit-Basic / Reddit-Advanced, Step~1 runs on the pre-processed thread (post body + top-3 comment chain with URLs, handles, and subreddit names stripped). Step~2 is identical to the public variants.

\smallskip
\textbf{Output for both steps is strict JSON; no surrounding text.}
\end{tcolorbox}

%% file: tables/appendix/cross_corpus_js.tex
\begin{table}[t]
\centering
\small
\setlength{\tabcolsep}{4.2pt}
\renewcommand{\arraystretch}{1.08}
\caption{
Cross-corpus JS between train splits. 
Behav-JS stays near the reference floor for same-source Basic/Advanced variants and is much larger across Public/Reddit populations; Struct-JS is more sensitive to surface-form changes.
}
\label{tab:cross_corpus_js}
\begin{tabular}{@{}lcccc@{}}
\toprule
& \textbf{Pub-B} & \textbf{Pub-A} & \textbf{Red-B} & \textbf{Red-A} \\
\midrule
\multicolumn{5}{@{}l}{\textit{Behav-JS, train$\leftrightarrow$train}} \\
Pub-B & ---   & 0.022 & 0.160 & 0.158 \\
Pub-A & 0.022 & ---   & 0.150 & 0.151 \\
Red-B & 0.160 & 0.150 & ---   & 0.022 \\
Red-A & 0.158 & 0.151 & 0.022 & ---   \\
\midrule
\multicolumn{5}{@{}l}{\textit{Struct-JS, train$\leftrightarrow$train}} \\
Pub-B & ---   & 0.313 & 0.716 & 0.739 \\
Pub-A & 0.313 & ---   & 0.718 & 0.699 \\
Red-B & 0.716 & 0.718 & ---   & 0.308 \\
Red-A & 0.739 & 0.699 & 0.308 & ---   \\
\midrule
\textit{Reference}
& \multicolumn{4}{c}{\textit{Behav 0.023--0.030; Struct 0.076--0.094}} \\
\bottomrule
\end{tabular}
\end{table}

%% file: tables/appendix/full_alignment_table.tex
\begin{table*}[!htbp]
\centering
\footnotesize
\setlength{\tabcolsep}{2.0pt}
\renewcommand{\arraystretch}{1.04}
\caption{
Full per-corpus distribution-alignment results under Claude Sonnet 4 labelling (JS divergence, $\downarrow$).
Average results are reported in Table~\ref{tab:main-eval}; this table expands the per-corpus structural and external checks.
Bold and underline mark best and second-best non-reference values in each column.
}
\label{tab:full-alignment}
\resizebox{\textwidth}{!}{%
\begin{tabular}{lcccccccccccccccc}
\toprule
& \multicolumn{4}{c}{\textbf{Public-Basic}}
& \multicolumn{4}{c}{\textbf{Public-Advanced}}
& \multicolumn{4}{c}{\textbf{Reddit-Basic}}
& \multicolumn{4}{c}{\textbf{Reddit-Advanced}} \\
\cmidrule(lr){2-5}
\cmidrule(lr){6-9}
\cmidrule(lr){10-13}
\cmidrule(lr){14-17}
\textbf{Method}
& Behav & Struct & Ext-Act & Ext-E/T
& Behav & Struct & Ext-Act & Ext-E/T
& Behav & Struct & Ext-Act & Ext-E/T
& Behav & Struct & Ext-Act & Ext-E/T \\
\midrule
AutoPAL
& 0.244 & 0.172 & 0.157 & 0.074
& 0.256 & \textbf{0.179} & 0.183 & 0.052
& 0.187 & \textbf{0.148} & 0.163 & 0.084
& 0.274 & 0.182 & 0.165 & 0.081 \\

ConceptPersona
& 0.236 & \textbf{0.162} & \underline{0.137} & 0.083
& \underline{0.252} & 0.200 & 0.163 & 0.053
& 0.181 & 0.164 & 0.172 & \textbf{0.063}
& \underline{0.269} & 0.182 & 0.144 & 0.082 \\

DiaSynth
& 0.247 & 0.169 & 0.138 & 0.081
& 0.259 & 0.211 & 0.161 & 0.053
& \underline{0.162} & 0.156 & 0.171 & 0.069
& 0.288 & 0.169 & 0.152 & \underline{0.075} \\

FaithfulPersona
& 0.235 & \underline{0.164} & 0.140 & \textbf{0.069}
& 0.263 & 0.188 & \underline{0.153} & 0.053
& \textbf{0.156} & \underline{0.150} & \underline{0.146} & 0.069
& 0.280 & 0.176 & \underline{0.140} & 0.095 \\

ICL
& 0.327 & 0.250 & 0.238 & \underline{0.071}
& 0.323 & 0.215 & 0.264 & 0.067
& 0.310 & 0.238 & 0.262 & 0.093
& 0.316 & 0.220 & 0.229 & 0.109 \\

Interlocutor
& \underline{0.232} & 0.175 & 0.171 & 0.089
& 0.254 & 0.190 & 0.209 & 0.074
& 0.265 & 0.197 & 0.178 & 0.080
& 0.284 & 0.169 & 0.188 & 0.080 \\

PersonaLens
& 0.245 & 0.175 & 0.159 & 0.076
& 0.271 & \underline{0.187} & 0.182 & \underline{0.047}
& 0.210 & 0.176 & 0.170 & \underline{0.067}
& 0.284 & \underline{0.166} & 0.160 & \underline{0.075} \\

\midrule
\textbf{GroupPersona (ours)}
& \textbf{0.148} & 0.177 & \textbf{0.115} & 0.076
& \textbf{0.228} & 0.194 & \textbf{0.120} & \textbf{0.043}
& \textbf{0.156} & 0.170 & \textbf{0.109} & 0.086
& \textbf{0.174} & \textbf{0.125} & \textbf{0.103} & \textbf{0.067} \\

\midrule
\textit{Reference (reference--reference)}
& \textit{0.030} & \textit{0.076} & --- & ---
& \textit{0.024} & \textit{0.087} & --- & ---
& \textit{0.029} & \textit{0.094} & --- & ---
& \textit{0.023} & \textit{0.086} & --- & --- \\
\bottomrule
\end{tabular}}
\end{table*}

%% file: tables/appendix/full_ablation_table.tex
\begin{table*}[!htbp]
\centering
\footnotesize
\setlength{\tabcolsep}{2.0pt}
\renewcommand{\arraystretch}{1.04}
\caption{
Full per-corpus GroupPersona ablation results under Claude Sonnet 4 labelling (JS divergence, $\downarrow$).
Average results are reported in Table~\ref{tab:ablation-full}; this table expands the per-corpus structural and external checks.
Lower is better.
}
\label{tab:ablation-full-appendix}
\resizebox{\textwidth}{!}{%
\begin{tabular}{lcccccccccccccccc}
\toprule
& \multicolumn{4}{c}{\textbf{Public-B}}
& \multicolumn{4}{c}{\textbf{Public-A}}
& \multicolumn{4}{c}{\textbf{Reddit-B}}
& \multicolumn{4}{c}{\textbf{Reddit-A}} \\
\cmidrule(lr){2-5}
\cmidrule(lr){6-9}
\cmidrule(lr){10-13}
\cmidrule(lr){14-17}
\textbf{Variant}
& Behav & Struct & Ext-Act & Ext-E/T
& Behav & Struct & Ext-Act & Ext-E/T
& Behav & Struct & Ext-Act & Ext-E/T
& Behav & Struct & Ext-Act & Ext-E/T \\
\midrule
\textbf{GroupPersona-full}
& 0.148 & 0.177 & 0.115 & 0.076
& 0.228 & 0.194 & 0.120 & 0.043
& 0.156 & 0.170 & 0.109 & 0.086
& 0.174 & 0.125 & 0.103 & 0.067 \\

\midrule
\multicolumn{17}{l}{\textit{Group-discovery ablations}} \\
\quad no LLM rule verifier
& 0.160 & 0.177 & 0.125 & 0.066
& 0.230 & 0.194 & 0.143 & 0.042
& 0.174 & 0.170 & 0.132 & 0.076
& 0.191 & 0.125 & 0.111 & 0.082 \\

\quad $k$-means groups
& 0.182 & 0.177 & 0.149 & 0.088
& 0.248 & 0.194 & 0.142 & 0.045
& 0.194 & 0.170 & 0.153 & 0.086
& 0.205 & 0.125 & 0.128 & 0.076 \\

\quad random groups
& 0.233 & 0.177 & 0.160 & 0.091
& 0.264 & 0.194 & 0.185 & 0.051
& 0.235 & 0.170 & 0.185 & 0.080
& 0.246 & 0.125 & 0.164 & 0.093 \\

\midrule
\multicolumn{17}{l}{\textit{Profile-conditioning ablations}} \\
\quad $-$ source prefix
& 0.129 & 0.126 & 0.127 & 0.072
& 0.265 & 0.197 & 0.155 & 0.051
& 0.231 & 0.219 & 0.166 & 0.082
& 0.176 & 0.165 & 0.137 & 0.067 \\

\quad $-$ behavior roots
& 0.223 & 0.191 & 0.153 & 0.057
& 0.281 & 0.229 & 0.185 & 0.057
& 0.190 & 0.209 & 0.169 & 0.093
& 0.178 & 0.148 & 0.158 & 0.080 \\

\quad $-$ behavior tendencies
& 0.149 & 0.140 & 0.127 & 0.072
& 0.265 & 0.204 & 0.152 & 0.046
& 0.174 & 0.161 & 0.143 & 0.063
& 0.155 & 0.110 & 0.117 & 0.094 \\

\quad $-$ structural statistics
& 0.256 & 0.232 & 0.169 & 0.069
& 0.276 & 0.238 & 0.191 & 0.031
& 0.216 & 0.210 & 0.184 & 0.086
& 0.224 & 0.204 & 0.146 & 0.080 \\
\bottomrule
\end{tabular}}
\end{table*}

%% file: tables/appendix/bootstrap_ci.tex
\begin{table}[t]
\centering
\small
\setlength{\tabcolsep}{3.5pt}
\caption{Conversation-level bootstrap 95\% CIs for Behav-JS (200 resamples). \emph{GA}: GroupPersona-full. \emph{Best BL}: best baseline per (corpus, family) cell (named in App.~\ref{app:cross_family}). $\Diamond$ marks (corpus, family) cells where the GA and best-baseline CIs are disjoint (separation at the 95\% level).}
\label{tab:bootstrap_ci}
\resizebox{\columnwidth}{!}{%
\begin{tabular}{llcc}
\toprule
\textbf{Corpus} & \textbf{Family} & \textbf{GA 95\% CI} & \textbf{Best BL 95\% CI} \\
\midrule
\multirow{3}{*}{Public-Basic}
 & Claude Sonnet 4    & [0.140, 0.166]$\Diamond$ & [0.223, 0.246] \\
 & DeepSeek-R1 & [0.149, 0.178]$\Diamond$ & [0.225, 0.246] \\
 & Llama-3     & [0.158, 0.185]$\Diamond$ & [0.223, 0.246] \\
\midrule
\multirow{3}{*}{Public-Advanced}
 & Claude Sonnet 4    & [0.213, 0.241]$\Diamond$ & [0.241, 0.264] \\
 & DeepSeek-R1 & [0.229, 0.260] & [0.243, 0.263] \\
 & Llama-3     & [0.236, 0.266] & [0.239, 0.263] \\
\midrule
\multirow{3}{*}{Reddit-Basic}
 & Claude Sonnet 4    & [0.147, 0.172] & [0.149, 0.170] \\
 & DeepSeek-R1 & [0.154, 0.178] & [0.161, 0.184] \\
 & Llama-3     & [0.168, 0.192] & [0.167, 0.192] \\
\midrule
\multirow{3}{*}{Reddit-Advanced}
 & Claude Sonnet 4    & [0.167, 0.191]$\Diamond$ & [0.260, 0.279] \\
 & DeepSeek-R1 & [0.168, 0.192]$\Diamond$ & [0.259, 0.283] \\
 & Llama-3     & [0.181, 0.206]$\Diamond$ & [0.266, 0.289] \\
\bottomrule
\end{tabular}}
\end{table}

%% file: tables/appendix/cross_family_pipeline.tex
\begin{table}[t]
\centering
\small
\setlength{\tabcolsep}{3pt}
\renewcommand{\arraystretch}{1.05}
\caption{Per-corpus Behav-JS for GroupPersona-full vs.\ ConceptPersona under five cross-family configurations. \emph{In-family}: same family synthesizes and labels. \emph{Out-of-family}: Claude Sonnet 4 synthesizes; another family re-labels the same records. $\Delta\%$ is GroupPersona's reduction over ConceptPersona on that cell. The two out-of-family rows show that DeepSeek-R1 and Llama-3, scoring the same Claude Sonnet 4-synthesised corpus, reproduce the GroupPersona ranking; on the two corpora with a large Claude Sonnet 4 gap (PB, RA), out-of-family Claude Sonnet 4$\to$Llama-3 actually retains a larger reduction than Llama-3 in-family.}
\label{tab:cross_family_full}
\resizebox{\columnwidth}{!}{%
\begin{tabular}{lcccc}
\toprule
\textbf{Setup} & Public-B & Public-A & Reddit-B & Reddit-A \\
\midrule
\multicolumn{5}{l}{\textit{In-family pipeline}} \\
Claude Sonnet 4    & $-37.2\%$ & $-9.7\%$  & $-13.7\%$ & $-35.4\%$ \\
DeepSeek-R1 & $-32.6\%$ & $-3.1\%$  & $-14.6\%$ & $-34.2\%$ \\
Llama-3     & $-28.0\%$ & $+0.7\%$  & $-10.9\%$ & $-30.1\%$ \\
\midrule
\multicolumn{5}{l}{\textit{Out-of-family (Claude Sonnet 4 synth, other labels)}} \\
Claude Sonnet 4$\to$DeepSeek-R1   & $-37.3\%$ & $-8.2\%$ & $-13.3\%$ & $-35.4\%$ \\
Claude Sonnet 4$\to$Llama-3       & $-34.4\%$ & $-6.7\%$ & $-10.0\%$ & $-33.4\%$ \\
\bottomrule
\end{tabular}}
\end{table}

%% file: tables/appendix/judge_role_table.tex
\begin{table}[t]
\centering
\small
\setlength{\tabcolsep}{4.5pt}
\renewcommand{\arraystretch}{1.10}
\caption{Human audit of the two LLM-driven steps. Label-Q rates per-dialogue behavior labels; Agreement rates rule-verification decisions. Scores are on a 1--5 Likert scale.}
\label{tab:judge-role}
\begin{tabular}{lcc}
\toprule
\textbf{Family}
 & \textbf{Label-Q.}\,$\uparrow$
 & \textbf{Agreement}\,$\uparrow$ \\
\midrule
Claude Sonnet 4    & $4.07 \pm 0.32$         & $\mathbf{3.83 \pm 0.12}$ \\
DeepSeek-R1 & $4.08 \pm 0.33$         & $3.76 \pm 0.14$         \\
Llama-3     & $\mathbf{4.11 \pm 0.32}$ & $3.70 \pm 0.24$         \\
\bottomrule
\end{tabular}
\end{table}

%% file: tables/appendix/group_diag_table.tex
\begin{table}[t]
\centering
\small
\setlength{\tabcolsep}{2.8pt}
\renewcommand{\arraystretch}{1.08}
\caption{
Group-discovery diagnostics and threshold selection.
Bold marks the locked configuration.
}
\label{tab:group-diagnostics}
\begin{tabular}{@{}lrrrrrr@{}}
\toprule
\multicolumn{7}{c}{\textbf{(a) Locked configuration: train summary}} \\
\midrule
\textbf{Corpus} & \textbf{Rules} & \textbf{Groups} & \textbf{Resid.} & \textbf{Drop} & \textbf{Root-4} & \textbf{Root-3} \\
\midrule
Pub-B &  88 &  96 & 10.2 & 1.07 & 87 & 9 \\
Pub-A & 109 & 104 &  8.6 & 1.17 & 96 & 8 \\
Red-B & 121 &  88 &  7.5 & 1.46 & 80 & 8 \\
Red-A & 142 &  73 &  9.4 & 1.67 & 65 & 8 \\
\midrule
\multicolumn{7}{c}{\textbf{(b) Held-out validation sweep, mean over corpora}} \\
\midrule
$\tau_h$ & $\tau_l$ & $\tau_j$ & $\tau_s$ & \textbf{Resid.} & \textbf{MedRoot} & \textbf{Val-JS} \\
\midrule
\textbf{0.60} & \textbf{1.15} & \textbf{0.50} & \textbf{3} & \textbf{8.9}  & \textbf{4} & \textbf{0.183} \\
0.60 & 1.15 & 0.45 & 3 & 5.4  & 4 & 0.188 \\
0.60 & 1.15 & 0.55 & 3 & 13.7 & 4 & 0.190 \\
0.60 & 1.20 & 0.50 & 3 & 9.3  & 4 & 0.187 \\
0.65 & 1.15 & 0.50 & 3 & 9.0  & 4 & 0.186 \\
0.65 & 1.20 & 0.50 & 5 & 17.1 & 4 & 0.204 \\
0.70 & 1.20 & 0.50 & 5 & 17.6 & 3 & 0.207 \\
0.70 & 1.30 & 0.50 & 5 & 18.0 & 3 & 0.211 \\
\bottomrule
\end{tabular}
\end{table}

%% file: tables/appendix/llm_pattern_judge_prompt.tex
\noindent\begin{tcolorbox}[breakable, colback=gray!5, colframe=gray!40, arc=2pt,
  left=4pt, right=4pt, top=4pt, bottom=4pt, fontupper=\small,
  title={\small\textbf{LLM pattern-judge prompt (full)}},
  coltitle=black, colbacktitle=gray!12,
  width=\linewidth]
\footnotesize

\textbf{Role.} You are a senior dialogue researcher reviewing one \emph{minimal} association rule for inclusion in a group-conditioned synthesis pipeline. The rule has been backward-pruned to the smallest pair-subset whose predictive power on the consequent is preserved (within $\delta{=}0.05$ confidence) relative to the parent rule. Accepted rules will be used to drop the consequent pair from a dialogue whenever the dialogue's other 11 pairs already contain the minimal antecedent. Your job is to decide whether the minimal rule still reflects a \emph{semantically} plausible user-behaviour regularity, not merely a statistical artefact, given its corpus statistics, the parent rule, and three example conversations that match the antecedent.

\smallskip
\textbf{Input.}
\begin{itemize}
\setlength{\itemsep}{1pt}
\item \texttt{minimal\_rule}: the minimal antecedent $A^\star$ and consequent $(D{=}d)$, e.g.\
``(\texttt{primary\_intent\_type=Chitchat}) AND (\texttt{user\_type=Casual}) $\Rightarrow$ (\texttt{user\_goal\_profile=Chatter})''.
\item \texttt{parent\_rule}: the rule from which $A^\star$ was pruned (or the same rule if already minimal), so you can sanity-check that minimisation did not strip a semantically necessary atom.
\item \texttt{stats}: \texttt{support}, \texttt{confidence}, \texttt{lift}, \texttt{pattern\_score} computed on $A^\star$.
\item \texttt{examples}: 3 example conversations (verbatim turns) whose 12 attribute-value pairs contain $A^\star$; for each example, also show whether the pair-set also contains the consequent.
\end{itemize}

\smallskip
\textbf{Decision criterion.}
The minimal rule is \emph{reasonable} iff
(i) a reader can articulate \emph{why} the minimal antecedent would imply the consequent (causal, definitional, or strong behavioural correlation) independent of the numbers,
(ii) the minimal antecedent is not missing a semantically necessary atom that the parent carried,
(iii) the statistics meet the pre-filter floors (\texttt{confidence} $\geq 0.80$, \texttt{lift} $\geq 1.3$, \texttt{support} $\geq 0.03$; near-threshold rules deserve extra scrutiny), \emph{and}
(iv) at least 2 of the 3 example conversations visibly contain both the antecedent and the consequent.

\smallskip
\textbf{Reject if any of the following.}
\begin{itemize}
\setlength{\itemsep}{1pt}
\item \emph{Sampling artefact}: lift barely above 1; consequent is the marginal mode anyway.
\item \emph{Tautology}: consequent is a rephrasing of the antecedent on the same dimension.
\item \emph{Example contradiction}: conversations matching the antecedent do not exhibit the consequent.
\item \emph{Generic-prior consequent}: consequent is \texttt{unknown} or a default category and the lift is driven by that prior.
\item \emph{Template-induced co-occurrence}: the link is mechanically forced by the corpus generation pipeline (e.g.\ ``\texttt{interaction\_mode=QA-style} $\Rightarrow$ \texttt{turn\_count=2}'' in a synthesised QA dataset).
\end{itemize}

\smallskip
\textbf{Confidence calibration.}
\begin{itemize}
\setlength{\itemsep}{1pt}
\item \emph{Accept, strong stats} (conf~$\geq 0.7$, lift~$\geq 2$) + all 3 examples support $\to$ output confidence in $[0.85, 0.99]$.
\item \emph{Accept, moderate stats} (conf~$0.5$--$0.7$) + 2/3 examples support $\to$ output confidence in $[0.65, 0.85]$.
\item \emph{Accept, borderline} (mixed evidence) $\to$ output confidence in $[0.45, 0.65]$.
\item \emph{Reject, weak stats} (conf~$<0.5$, lift~$<1.5$) + no examples support $\to$ output confidence in $[0.70, 0.95]$.
\item \emph{Reject, strong stats but tautological or artefact} + examples contradict $\to$ output confidence in $[0.55, 0.80]$.
\end{itemize}

\smallskip
\textbf{Output (strict JSON; no surrounding text).}
\begin{verbatim}
{
  "is_reasonable": true|false,
  "confidence": 0.0,
  "reasoning": "..."
}
\end{verbatim}

The \texttt{reasoning} field must be one short sentence that cites the concrete signal driving the decision (e.g.\ ``all three examples show short imperative turns, consistent with the consequent'').
\end{tcolorbox}

%% file: tables/appendix/residual_case_study_table.tex
\begin{table*}[t]
\centering
\footnotesize
\setlength{\tabcolsep}{4pt}
\renewcommand{\arraystretch}{1.10}
\caption{Public-Basic case study of two groups under the locked configuration $(\tau_{\mathrm{hom}}, \tau_{\mathrm{lift}}, \tau_{\mathrm{size}}, K_{\max}) = (0.60, 1.15, 3, 4)$. \textbf{Group~A} is the largest 4-root cluster; \textbf{Group~B} is the smallest 2-root cluster admitted under the size floor. Roots are the (dim, value) pairs that satisfy in-cluster homogeneity $\geq \tau_{\mathrm{hom}}$, lift $\geq \tau_{\mathrm{lift}}$, and LLM-judge endorsement (\S\ref{sec:root-selection}); non-root rows show the conditional marginal $\Pr[D' \mid g]$ for the corresponding dimension, which acts as a \emph{characteristic pattern} at synthesis time.}
\label{tab:residual_case_study}
\resizebox{\textwidth}{!}{%
\begin{tabular}{lll}
\toprule
\textbf{Field} & \textbf{Group~A (casual chitchat)} & \textbf{Group~B (one-shot impatient)} \\
\midrule
Cluster size           & 53 records & 5 records \\
\# of roots admitted   & 4          & 2          \\
\midrule
\multicolumn{3}{l}{\textbf{Root signature} --- the group identifier} \\
\midrule
primary\_intent\_type      & \textbf{Chitchat} (root) & non-root: Info-seeking 60\%, Music 20\%, Smart\_Home 20\% \\
interaction\_mode          & \textbf{Multi-turn\_Dialog-style} (root) & non-root: QA-style 100\% \\
user\_type                 & \textbf{Casual} (root) & \textbf{One-shot\_User} (root) \\
user\_goal\_profile        & \textbf{Chatter} (root) & non-root: Explorer 60\%, Goal-switcher 20\%, Tasker 20\% \\
response\_brevity          & non-root: Long 81\%, Medium 11\%, Short 8\% & \textbf{Short} (root) \\
\midrule
\multicolumn{3}{l}{\textbf{Non-root marginals} --- characteristic patterns at synthesis} \\
\midrule
persistence\_level         & Low 53\%, Medium 34\%, High 13\% & Medium 80\%, High 20\% \\
topic\_cohesion            & Medium 77\%, Fragmented 17\%, High 6\% & High 80\%, Fragmented 20\% \\
politeness\_strategy       & Formal 36\%, Polite 25\%, Friendly 23\% & Direct 60\%, Polite 40\% \\
interaction\_flexibility   & Adaptive 74\%, Static 26\% & Static 80\%, Adaptive 20\% \\
persona\_adaptability      & High 83\%, Low 17\% & Low 60\%, High 40\% \\
repair\_behavior           & Clarify 47\%, Rephrase 23\%, Ignore 19\% & Rephrase 80\%, Retry 20\% \\
error\_recovery\_style     & Restart 45\%, Clarification\_Req.\ 23\%, Task\_Abandon.\ 17\% & Repetition 80\%, Restart 20\% \\
\bottomrule
\end{tabular}}
\end{table*}

%% file: tables/appendix/metadata_enrichment_prompt.tex
\noindent\begin{tcolorbox}[breakable, colback=gray!5, colframe=gray!40, arc=2pt,
  left=4pt, right=4pt, top=4pt, bottom=4pt, fontupper=\small,
  title={\small\textbf{Group metadata enrichment prompt}},
  coltitle=black, colbacktitle=gray!12,
  width=\linewidth]
\footnotesize

This prompt produces the three conditioning ingredients consumed by the user agent at synthesis time --- (ii) group\_persona, (iii) conversation\_patterns, (iv) group\_statistics --- from one cluster of reference dialogues. The fidelity of this output depends directly on within-cluster homogeneity: a cluster whose records disagree on most dimensions yields a generic, contradictory persona; a cluster whose records share LLM-verified roots yields a sharp, behaviourally-anchored persona.

\smallskip
\textbf{Role.} You are a conversation analyst. Given a set of reference dialogues assigned to the same behavioural group $g$, together with $g$'s root signature (a small tuple of (dimension, value) pairs shared by the cluster's records), produce a compact metadata object that conditions downstream synthesis.

\smallskip
\textbf{Input.}
(a) Root signature for $g$: e.g.\ \texttt{\{user\_type=Casual, primary\_intent=Chitchat, user\_goal=Chatter, interaction\_mode=Multi-turn\_Dialog-style\}}.
(b) Up to $n \le 30$ reference dialogues belonging to $g$ (full text + per-record 12-dim labels + structural fields).

\smallskip
\textbf{Output (strict JSON, three fields).}
\begin{itemize}
\setlength{\itemsep}{0pt}
\item \texttt{group\_persona} (2--3 sentences) --- typical goals, tone, and interaction style of users in $g$. Anchor every claim to evidence in the provided dialogues. Do not invent demographics.
\item \texttt{conversation\_patterns} (3--5 bullets) --- canonical discourse flows, intent transitions, and the top-3 most informative \emph{non-root} dimensions with their in-cluster marginal (e.g.\ ``\texttt{response\_brevity}: Long~81\%, Medium~11\%, Short~8\%''). Patterns are transition-style, not content-specific.
\item \texttt{group\_statistics} --- numeric summary of the structural fields (mean and standard deviation of turn count, user/full word count, user/full utterance length). Computed on the provided dialogues, not estimated.
\end{itemize}

\smallskip
\textbf{Guidelines.}
Prefer concrete observed behaviours over abstract descriptors.
For high-homogeneity clusters (top-4 dim agreement $\geq 0.75$), the persona should foreground the root dimensions verbatim;
for lower-homogeneity clusters, mark unstable dimensions with their marginal (``mostly Casual; sometimes One-shot users'').
Never claim a behaviour that does not appear in $\geq 1/3$ of the provided dialogues.

\smallskip
\textbf{How cluster quality propagates.} The same prompt is run on every clustering scheme:
\emph{GroupPersona} provides clusters whose roots are LLM-verified and high-lift, so the persona is anchored to interpretable identifiers;
\emph{K-means} on one-hot labels provides geometrically tight clusters whose modal values have no semantic verification and may be corpus artefacts;
\emph{Random clusters} provide clusters whose within-cluster modal values are essentially the corpus marginal, so the persona reduces to a generic description with weak conditioning power.
The cluster-quality ablations in \S\ref{sec:exp-q2} measure this propagation.
\end{tcolorbox}

%% file: tables/appendix/gt_calibration_full_table.tex
\begin{table*}[t]
\centering
\footnotesize
\setlength{\tabcolsep}{2.4pt}
\renewcommand{\arraystretch}{1.00}
\caption{Full quality calibration to reference conversations. Scores use an 8-dimension LLM-as-Judge rubric; MAD is mean absolute deviation from the reference-conversation profile.}
\label{tab:gt-calibration-full}
\begin{tabular}{lcccccccc|c}
\toprule
\textbf{Method} & \textbf{Flow} & \textbf{H-Con} & \textbf{A-Con} & \textbf{Ctx} & \textbf{Turn} & \textbf{Topic} & \textbf{Use} & \textbf{Ovrl} & \textbf{MAD$\downarrow$} \\
\midrule
\textit{Reference} & \textit{6.66} & \textit{7.80} & \textit{6.42} & \textit{6.14} & \textit{7.58} & \textit{6.86} & \textit{7.28} & \textit{6.66} & \textit{0.00} \\
\midrule
AutoPAL          & 4.80 & 7.08 & 3.84 & 4.10 & 6.52 & 6.04 & 3.76 & 4.12 & 1.89 \\
ConceptPersona   & 8.24 & 8.32 & 8.10 & 8.34 & 8.48 & 8.74 & 8.16 & 8.10 & 1.39 \\
DiaSynth         & 8.64 & 8.96 & 8.52 & 8.58 & 8.66 & 9.12 & 8.48 & 8.56 & 1.77 \\
FaithfulPersona  & 8.10 & 8.38 & 7.92 & 7.86 & 8.24 & 8.30 & 7.92 & 7.84 & 1.15 \\
ICL              & 6.02 & 6.72 & 5.74 & 5.68 & 6.36 & 6.18 & 5.62 & 5.78 & \underline{0.91} \\
Interlocutor     & 5.58 & 6.00 & 5.10 & 5.16 & 5.88 & 5.70 & 4.78 & 5.30 & 1.49 \\
PersonaLens      & 8.44 & 8.78 & 8.58 & 8.24 & 8.48 & 8.66 & 8.32 & 8.42 & 1.57 \\
\midrule
\textbf{GroupPersona} & 7.34 & 8.60 & 6.22 & 7.18 & 8.20 & 7.96 & 7.06 & 7.02 & \textbf{0.63} \\
\bottomrule
\end{tabular}
\end{table*}

%% file: tables/appendix/conversation_synthesis_prompt.tex
\noindent\begin{tcolorbox}[breakable, colback=gray!5, colframe=gray!40, arc=2pt,
  left=6pt, right=6pt, top=4pt, bottom=4pt, fontupper=\small,
  title={\small\textbf{Group-conditioned multi-agent synthesis prompt}},
  coltitle=black, colbacktitle=gray!12,
  width=\linewidth]
\footnotesize

A two-agent loop synthesises one dialogue per group $g$: a \emph{user agent} conditioned on $g$'s metadata, and a \emph{fixed assistant agent} held identical across methods and ablations.

\medskip
\noindent\textbf{User-agent system prompt} (full configuration; ablations remove one bracketed clause, see below).

\smallskip
\begin{tcolorbox}[colback=white, colframe=gray!50, arc=1pt,
   left=4pt, right=4pt, top=3pt, bottom=3pt,
   fontupper=\scriptsize\ttfamily, breakable]
You are a user in a population that interacts with a voice assistant.\\
Your behaviour comes from the behavioural group described below; stay in character throughout the conversation.

\medskip
\textit{[root]}\ \ Group identifier: \{root\_dim$_1$\}=\{root\_val$_1$\}, \ldots, \{root\_dim$_4$\}=\{root\_val$_4$\}.

\textit{[persona]}\ \ \{group\_persona\_paragraph\}

\textit{[patterns]}\ \ Characteristic patterns (top-3 non-root marginals): \{dim$_a$=val\%,\ \ldots; dim$_b$=val\%,\ \ldots; dim$_c$=val\%,\ \ldots\}.

\textit{[stats]}\ \ Typical turn count $\mu \pm \sigma$, user-utterance length $\mu \pm \sigma$, word frequencies as in training.

\textit{[history]}\ \ Conversation seed (reference prefix): \{history\_text\}. Continue as the user, matching the seed's style.

\medskip
Output: only the next user turn; no role token. Stop when the conversation is naturally finished or the typical turn count is reached.
\end{tcolorbox}

\smallskip
\noindent\textbf{Assistant-agent system prompt} (identical across all methods).
\begin{tcolorbox}[colback=white, colframe=gray!50, arc=1pt,
   left=4pt, right=4pt, top=3pt, bottom=3pt,
   fontupper=\scriptsize\ttfamily, breakable]
You are a generic voice assistant. Be helpful, concise, and grounded in the conversation. You do not access any external metadata about the user.
\end{tcolorbox}

\smallskip
\noindent\textbf{Ablation mapping} (\S\ref{sec:exp-q3}). 
Conditioning-component ablations remove exactly one synthesis input from the user-agent prompt: 
\texttt{no\_history} drops \textit{[history]} ($-$ source prefix); 
\texttt{no\_persona} drops \textit{[persona]} ($-$ behavior roots, implemented as the natural-language root-guidance paragraph); 
\texttt{no\_patterns} drops \textit{[patterns]} ($-$ behavior tendencies); 
and \texttt{no\_stats} drops \textit{[stats]} ($-$ structural statistics).
The minimal \textit{[root]} identifier is retained in all four conditioning ablations to keep group selection fixed; it is used only as the selected group index and is not counted as a removable profile field. 
Cluster-quality ablations (\texttt{random}, \texttt{k-means}, \S\ref{sec:exp-q2}) replace the GroupPersona group identifier in \textit{[root]} with a random or $k$-means cluster id and re-run metadata enrichment (App.~\ref{app:metadata_prompt}) on the new cluster's records, so \textit{[persona]}, \textit{[patterns]}, and \textit{[stats]} carry weaker group information.
\end{tcolorbox}

%% file: tables/appendix/quality_calibration_prompt.tex
\noindent\begin{tcolorbox}[breakable, colback=gray!5, colframe=gray!40, arc=2pt,
  left=4pt, right=4pt, top=4pt, bottom=4pt, fontupper=\small,
  title={\small\textbf{LLM-as-Judge prompt for reference-conversation-anchored quality rating}},
  coltitle=black, colbacktitle=gray!12,
  width=\linewidth]
\footnotesize

\textbf{Role.} You are a senior dialogue-systems researcher rating one candidate conversation against a target reference corpus. The objective is to quantify how closely the candidate matches realistic dialogue from this corpus on standard conversation-quality dimensions, anchored against reference-corpus examples rather than against an idealised standard.

\smallskip
\textbf{Input.}
\begin{itemize}
\setlength{\itemsep}{1pt}
\item \texttt{candidate\_conversation}: the full multi-turn transcript to rate.
\item \texttt{corpus\_description}: a one-paragraph description of the reference corpus (domain, register, expected user behavior, interaction style).
\item \texttt{reference\_examples}: three reference conversations randomly sampled from the reference corpus, included verbatim so the judge can anchor against ``what the reference corpus looks like'' rather than an idealised prior.
\end{itemize}

\smallskip
\textbf{Calibration note in the prompt.}
The prompt explicitly states that reference conversations are imperfect --- they contain hesitation, mid-stream topic changes, false starts, repair, persistence, mood shifts, and varied turn lengths --- and that a candidate exhibiting none of this is \emph{further} from realism, not closer. The judge is instructed to anchor against the reference examples, not against an idealised standard.

\smallskip
\textbf{Dimensions (1--10, higher = more realistic match).}
\begin{itemize}
\setlength{\itemsep}{1pt}
\item \textbf{Flow}: whether the conversation progresses with natural pacing and a plausible arc.
\item \textbf{H-Con (User Consistency)}: whether the user maintains a coherent behavior pattern (verbosity, politeness, persistence) within the conversation.
\item \textbf{A-Con (Assistant Consistency)}: whether the assistant's style and capability stay consistent.
\item \textbf{Ctx (Context Utilisation)}: whether later turns demonstrably use information from earlier turns.
\item \textbf{Turn}: whether turn-length distribution and turn-taking rhythm match the reference examples.
\item \textbf{Topic}: whether topic introductions, maintenance, and transitions are plausible for this corpus.
\item \textbf{Use (Usefulness)}: whether the conversation reaches an outcome whose realism level matches the reference examples (\emph{not} whether the outcome is maximally informative).
\item \textbf{Ovrl (Overall Quality)}: a holistic assessment of how likely the conversation could be a reference sample from the reference corpus.
\end{itemize}

\smallskip
\textbf{Output (strict JSON).}
\begin{verbatim}
{
  "flow":   <float 1.0-10.0>,
  "h_con":  <float 1.0-10.0>,
  "a_con":  <float 1.0-10.0>,
  "ctx":    <float 1.0-10.0>,
  "turn":   <float 1.0-10.0>,
  "topic":  <float 1.0-10.0>,
  "use":    <float 1.0-10.0>,
  "ovrl":   <float 1.0-10.0>,
  "rationale": "..."
}
\end{verbatim}
The rationale field cites the concrete signal driving the lowest-scoring dimension.

\smallskip
\textbf{Aggregation.} For each (method, corpus) cell we score $500$ generated conversations and average per-dimension to obtain that cell's row in Table~\ref{tab:gt-calibration}. The reference-conversation row uses the same rubric applied to reference-corpus conversations. Mean absolute deviation (MAD) from the reference-conversation row is then computed across the 8 dimensions and averaged over the four corpora.
\end{tcolorbox}

%% file: tables/appendix/gt_calibration_per_corpus.tex
\begin{table}[!htbp]
\centering
\small
\setlength{\tabcolsep}{5.5pt}
\renewcommand{\arraystretch}{1.10}
\caption{
Per-corpus quality calibration to reference conversations.
Values are MAD from the reference-conversation quality profile across 8 LLM-as-Judge dimensions ($\downarrow$).
Bold and underline mark best and second-best non-reference values.
}
\label{tab:gt-calibration-per-corpus}
\begin{tabular}{@{}lccccc@{}}
\toprule
\textbf{Method}
& \textbf{Pub-B}
& \textbf{Pub-A}
& \textbf{Red-B}
& \textbf{Red-A}
& \textbf{Avg.} \\
\midrule
AutoPAL         & 1.76 & 1.88 & 1.93 & 2.00 & 1.89 \\
ConceptPersona  & 1.21 & 1.41 & 1.39 & 1.53 & 1.39 \\
DiaSynth        & 1.54 & 1.78 & 1.75 & 1.99 & 1.77 \\
FaithfulPersona & 1.07 & 1.08 & 1.16 & 1.28 & 1.15 \\
ICL             & \underline{0.74} & \underline{0.86} & \underline{1.02} & \underline{1.03} & \underline{0.91} \\
Interlocutor    & 1.43 & 1.51 & 1.49 & 1.53 & 1.49 \\
PersonaLens     & 1.38 & 1.58 & 1.56 & 1.75 & 1.57 \\
\midrule
\textbf{GroupPersona}
& \textbf{0.57} & \textbf{0.62} & \textbf{0.63} & \textbf{0.69} & \textbf{0.63} \\
\bottomrule
\end{tabular}
\end{table}

%% file: acl_latex.bbl
\begin{thebibliography}{35}
\providecommand{\natexlab}[1]{#1}

\bibitem[{Agrawal et~al.(1993)Agrawal, Imielinski, and
  Swami}]{agrawal1993mining}
Rakesh Agrawal, Tomasz Imielinski, and Arun Swami. 1993.
\newblock Mining association rules between sets of items in large databases.
\newblock In \emph{Proceedings of the 1993 ACM SIGMOD International Conference
  on Management of Data}, pages 207--216.

\bibitem[{Agrawal and Srikant(1994)}]{agrawal1994fast}
Rakesh Agrawal and Ramakrishnan Srikant. 1994.
\newblock Fast algorithms for mining association rules.
\newblock In \emph{Proceedings of the 20th International Conference on Very
  Large Data Bases}, pages 487--499.

\bibitem[{Baumgartner et~al.(2020)Baumgartner, Zannettou, Keegan, Squire, and
  Blackburn}]{baumgartner2020pushshift}
Jason Baumgartner, Savvas Zannettou, Brian Keegan, Megan Squire, and Jeremy
  Blackburn. 2020.
\newblock \href {https://doi.org/10.1609/icwsm.v14i1.7347} {{The Pushshift
  Reddit Dataset}}.
\newblock In \emph{Proceedings of the International AAAI Conference on Web and
  Social Media}, volume~14, pages 830--839.

\bibitem[{Bohus(2004)}]{bohus2004error}
Dan Bohus. 2004.
\newblock {Error Awareness and Recovery in Task-Oriented Spoken Dialogue
  Systems}.
\newblock Ph.d. thesis proposal, Carnegie Mellon University, Pittsburgh, PA.

\bibitem[{Bohus and Rudnicky(2005)}]{bohus2005sorry}
Dan Bohus and Alexander~I. Rudnicky. 2005.
\newblock \href {https://doi.org/10.18653/v1/2005.sigdial-1.14} {{Sorry and I
  Didn't Catch That! - An Investigation of Non-understanding Errors and
  Recovery Strategies}}.
\newblock In \emph{Proceedings of the 6th SIGdial Workshop on Discourse and
  Dialogue}, pages 128--143, Lisbon, Portugal. Special Interest Group on
  Discourse and Dialogue (SIGdial).

\bibitem[{Brown et~al.(2020)Brown, Mann, Ryder, Subbiah, Kaplan, Dhariwal,
  Neelakantan, Shyam, Sastry, Askell et~al.}]{brown2020language}
Tom~B Brown, Benjamin Mann, Nick Ryder, Melanie Subbiah, Jared Kaplan, Prafulla
  Dhariwal, Arvind Neelakantan, Pranav Shyam, Girish Sastry, Amanda Askell,
  et~al. 2020.
\newblock Language models are few-shot learners.
\newblock \emph{Advances in Neural Information Processing Systems}.

\bibitem[{Cheng et~al.(2024)Cheng, Liu, Xu, Hou, Ouyang, Leong, Li, Wu, and
  Zheng}]{cheng2024autopal}
Yi~Cheng, Wenge Liu, Kaishuai Xu, Wenjun Hou, Yi~Ouyang, Chak~Tou Leong, Wenjie
  Li, Xian Wu, and Yefeng Zheng. 2024.
\newblock Autopal: Autonomous adaptation to users for personal ai
  companionship.
\newblock \emph{arXiv preprint arXiv:2406.13960}.

\bibitem[{Dinan et~al.(2019)Dinan, Roller, Shuster, Fan, Auli, and
  Weston}]{dinan2018wizard}
Emily Dinan, Stephen Roller, Kurt Shuster, Angela Fan, Michael Auli, and Jason
  Weston. 2019.
\newblock \href {https://openreview.net/forum?id=r1l73iRqKm} {{Wizard of
  Wikipedia: Knowledge-Powered Conversational Agents}}.
\newblock In \emph{International Conference on Learning Representations}.

\bibitem[{Dou et~al.(2025)Dou, Galley, Peng, Kedzie, Cai, Ritter, Quirk, Xu,
  and Gao}]{dou-etal-2025-simulatorarena}
Yao Dou, Michel Galley, Baolin Peng, Chris Kedzie, Weixin Cai, Alan Ritter,
  Chris Quirk, Wei Xu, and Jianfeng Gao. 2025.
\newblock \href {https://doi.org/10.18653/v1/2025.emnlp-main.1786}
  {{S}imulator{A}rena: Are user simulators reliable proxies for multi-turn
  evaluation of {AI} assistants?}
\newblock In \emph{Proceedings of the 2025 Conference on Empirical Methods in
  Natural Language Processing}, pages 35212--35290, Suzhou, China. Association
  for Computational Linguistics.

\bibitem[{Finch and Choi(2020)}]{finch2020unified}
Sarah~E. Finch and Jinho~D. Choi. 2020.
\newblock \href {https://doi.org/10.18653/v1/2020.sigdial-1.29} {{Towards
  Unified Dialogue System Evaluation: A Comprehensive Analysis of Current
  Evaluation Protocols}}.
\newblock In \emph{Proceedings of the 21st Annual Meeting of the Special
  Interest Group on Discourse and Dialogue}, pages 236--245, 1st virtual
  meeting. Association for Computational Linguistics.

\bibitem[{Gu et~al.(2024)Gu, Jiang, Shi, Tan, Zhai, Xu, Li, Shen, Ma, Liu
  et~al.}]{gu2024survey}
Jiawei Gu, Xuhui Jiang, Zhichao Shi, Hexiang Tan, Xuehao Zhai, Chengjin Xu, Wei
  Li, Yinghan Shen, Shengjie Ma, Honghao Liu, et~al. 2024.
\newblock A survey on {LLM}-as-a-judge.
\newblock \emph{arXiv preprint arXiv:2411.15594}.

\bibitem[{Han et~al.(2007)Han, Cheng, Xin, and Yan}]{han2007frequent}
Jiawei Han, Hong Cheng, Dong Xin, and Xifeng Yan. 2007.
\newblock Frequent pattern mining: current status and future directions.
\newblock \emph{Data Mining and Knowledge Discovery}, 15(1):55--86.

\bibitem[{Hashimoto et~al.(2019)Hashimoto, Zhang, and
  Liang}]{hashimoto2018unifying}
Tatsunori~B. Hashimoto, Hugh Zhang, and Percy Liang. 2019.
\newblock \href {https://doi.org/10.18653/v1/N19-1169} {{Unifying Human and
  Statistical Evaluation for Natural Language Generation}}.
\newblock In \emph{Proceedings of the 2019 Conference of the North American
  Chapter of the Association for Computational Linguistics: Human Language
  Technologies, Volume 1 (Long and Short Papers)}, pages 1689--1701,
  Minneapolis, Minnesota. Association for Computational Linguistics.

\bibitem[{Holtzman et~al.(2020)Holtzman, Buys, Du, Forbes, and
  Choi}]{holtzman2020nucleus}
Ari Holtzman, Jan Buys, Li~Du, Maxwell Forbes, and Yejin Choi. 2020.
\newblock The curious case of neural text degeneration.
\newblock In \emph{Proceedings of ICLR}.

\bibitem[{Jandaghi et~al.(2024)Jandaghi, Sheng, Bai, Pujara, and
  Sidahmed}]{jandaghi2024faithful}
Pegah Jandaghi, Xianghai Sheng, Xinyi Bai, Jay Pujara, and Hakim Sidahmed.
  2024.
\newblock \href {https://doi.org/10.18653/v1/2024.nlp4convai-1.8} {{Faithful
  Persona-based Conversational Dataset Generation with Large Language Models}}.
\newblock In \emph{Proceedings of the 6th Workshop on NLP for Conversational AI
  (NLP4ConvAI 2024)}, pages 114--139, Bangkok, Thailand. Association for
  Computational Linguistics.

\bibitem[{Kim et~al.(2023)Kim, Ahn, Lee, Kim, Lee, Shin, and
  Lee}]{kim2023concept}
Donghyun Kim, Youbin Ahn, Chanhee Lee, Wongyu Kim, Kyong-Ho Lee, Donghoon Shin,
  and Yeonsoo Lee. 2023.
\newblock \href {https://doi.org/10.18653/v1/2023.eacl-main.252}
  {{Concept-based Persona Expansion for Improving Diversity of Persona-Grounded
  Dialogue}}.
\newblock In \emph{Proceedings of the 17th Conference of the European Chapter
  of the Association for Computational Linguistics}, pages 3471--3481,
  Dubrovnik, Croatia. Association for Computational Linguistics.

\bibitem[{Li et~al.(2016{\natexlab{a}})Li, Galley, Brockett, Gao, and
  Dolan}]{li2016diversity}
Jiwei Li, Michel Galley, Chris Brockett, Jianfeng Gao, and Bill Dolan.
  2016{\natexlab{a}}.
\newblock \href {https://doi.org/10.18653/v1/N16-1014} {{A Diversity-Promoting
  Objective Function for Neural Conversation Models}}.
\newblock In \emph{Proceedings of the 2016 Conference of the North American
  Chapter of the Association for Computational Linguistics: Human Language
  Technologies}, pages 110--119, San Diego, California. Association for
  Computational Linguistics.

\bibitem[{Li et~al.(2016{\natexlab{b}})Li, Galley, Brockett, Spithourakis, Gao,
  and Dolan}]{li2016persona}
Jiwei Li, Michel Galley, Chris Brockett, Georgios Spithourakis, Jianfeng Gao,
  and Bill Dolan. 2016{\natexlab{b}}.
\newblock \href {https://doi.org/10.18653/v1/P16-1094} {{A Persona-Based Neural
  Conversation Model}}.
\newblock In \emph{Proceedings of the 54th Annual Meeting of the Association
  for Computational Linguistics (Volume 1: Long Papers)}, pages 994--1003,
  Berlin, Germany. Association for Computational Linguistics.

\bibitem[{Li et~al.(2025{\natexlab{a}})Li, Liu, Liu, Ye, Jiang, Zhang, and
  Ruan}]{li2025mads}
Mingjin Li, Yu~Liu, Huayi Liu, Xiang Ye, Chao Jiang, Hongguang Zhang, and
  Yu~Ruan. 2025{\natexlab{a}}.
\newblock \href {https://doi.org/10.18653/v1/2025.emnlp-industry.26} {{MADS:
  Multi-Agent Dialogue Simulation for Diverse Persuasion Data Generation}}.
\newblock In \emph{Proceedings of the 2025 Conference on Empirical Methods in
  Natural Language Processing: Industry Track}, pages 399--415, Suzhou, China.
  Association for Computational Linguistics.

\bibitem[{Li et~al.(2025{\natexlab{b}})Li, Bantupalli, Dharmani, Zhang, and
  Shang}]{li2025toward}
Xintong Li, Jalend Bantupalli, Ria Dharmani, Yuwei Zhang, and Jingbo Shang.
  2025{\natexlab{b}}.
\newblock \href {https://doi.org/10.18653/v1/2025.emnlp-main.580} {{Toward
  Multi-Session Personalized Conversation: A Large-Scale Dataset and
  Hierarchical Tree Framework for Implicit Reasoning}}.
\newblock In \emph{Proceedings of the 2025 Conference on Empirical Methods in
  Natural Language Processing}, pages 11493--11506, Suzhou, China. Association
  for Computational Linguistics.

\bibitem[{Li et~al.(2017)Li, Su, Shen, Li, Cao, and Niu}]{li2017dailydialog}
Yanran Li, Hui Su, Xiaoyu Shen, Wenjie Li, Ziqiang Cao, and Shuzi Niu. 2017.
\newblock \href {https://aclanthology.org/I17-1099/} {{DailyDialog: A Manually
  Labelled Multi-turn Dialogue Dataset}}.
\newblock In \emph{Proceedings of the Eighth International Joint Conference on
  Natural Language Processing (Volume 1: Long Papers)}, pages 986--995, Taipei,
  Taiwan. Asian Federation of Natural Language Processing.

\bibitem[{Mazar{'e} et~al.(2018)Mazar{'e}, Humeau, Raison, and
  Bordes}]{mazare2018training}
Pierre-Emmanuel Mazar{'e}, Samuel Humeau, Martin Raison, and Antoine Bordes.
  2018.
\newblock \href {https://doi.org/10.18653/v1/D18-1298} {{Training Millions of
  Personalized Dialogue Agents}}.
\newblock In \emph{Proceedings of the 2018 Conference on Empirical Methods in
  Natural Language Processing}, pages 2775--2779, Brussels, Belgium.
  Association for Computational Linguistics.

\bibitem[{Mehri and Eskenazi(2020)}]{mehri2020usr}
Shikib Mehri and Maxine Eskenazi. 2020.
\newblock \href {https://doi.org/10.18653/v1/2020.acl-main.64} {{USR: An
  Unsupervised and Reference Free Evaluation Metric for Dialog Generation}}.
\newblock In \emph{Proceedings of the 58th Annual Meeting of the Association
  for Computational Linguistics}, pages 681--707, Online. Association for
  Computational Linguistics.

\bibitem[{Occhipinti et~al.(2025)Occhipinti, Guerini, and
  Nissim}]{occhipinti2025harry}
Daniela Occhipinti, Marco Guerini, and Malvina Nissim. 2025.
\newblock \href {https://doi.org/10.18653/v1/2025.acl-long.879} {{When Harry
  Meets Superman: The Role of The Interlocutor in Persona-Based Dialogue
  Generation}}.
\newblock In \emph{Proceedings of the 63rd Annual Meeting of the Association
  for Computational Linguistics (Volume 1: Long Papers)}, pages 17964--17985,
  Vienna, Austria. Association for Computational Linguistics.

\bibitem[{Sarikaya(2017)}]{sarikaya2017technology}
Ruhi Sarikaya. 2017.
\newblock \href {https://doi.org/10.1109/MSP.2016.2617341} {{The Technology
  Behind Personal Digital Assistants: An Overview of the System Architecture
  and Key Components}}.
\newblock \emph{IEEE Signal Processing Magazine}, 34(1):67--81.

\bibitem[{Suresh et~al.(2025)Suresh, Mengjun, Pranav, and
  Chng}]{suresh2025diasynth}
Sathya~Krishnan Suresh, Wu~Mengjun, Tushar Pranav, and Eng~Siong Chng. 2025.
\newblock \href {https://doi.org/10.18653/v1/2025.findings-naacl.40}
  {{DiaSynth: Synthetic Dialogue Generation Framework for Low Resource Dialogue
  Applications}}.
\newblock In \emph{Findings of the Association for Computational Linguistics:
  NAACL 2025}, pages 673--690, Albuquerque, New Mexico. Association for
  Computational Linguistics.

\bibitem[{Wang et~al.(2025)Wang, Li, Yang, Zhou, Jiang, and Li}]{wang2025know}
Kuang Wang, Xianfei Li, Shenghao Yang, Li~Zhou, Feng Jiang, and Haizhou Li.
  2025.
\newblock \href {https://doi.org/10.18653/v1/2025.acl-long.1025} {{Know You
  First and Be You Better: Modeling Human-Like User Simulators via Implicit
  Profiles}}.
\newblock In \emph{Proceedings of the 63rd Annual Meeting of the Association
  for Computational Linguistics (Volume 1: Long Papers)}, pages 21082--21107,
  Vienna, Austria. Association for Computational Linguistics.

\bibitem[{Welleck et~al.(2020)Welleck, Kulikov, Roller, Dinan, Cho, and
  Weston}]{welleck2020neural}
Sean Welleck, Ilia Kulikov, Stephen Roller, Emily Dinan, Kyunghyun Cho, and
  Jason Weston. 2020.
\newblock Neural text generation with unlikelihood training.
\newblock In \emph{Proceedings of ICLR}.

\bibitem[{Wolf et~al.(2019)Wolf, Sanh, Chaumond, and
  Delangue}]{wolf2019transfertransfo}
Thomas Wolf, Victor Sanh, Julien Chaumond, and Clement Delangue. 2019.
\newblock Transfertransfo: A transfer learning approach for neural network
  based conversational agents.
\newblock \emph{arXiv preprint arXiv:1901.08149}.

\bibitem[{Yang et~al.(2025)Yang, Liu, Xiao, Zhao, Tang, Li, Yuan, Yang, and
  Lin}]{yang2025crafting}
Bohao Yang, Dong Liu, Chenghao Xiao, Kun Zhao, Chen Tang, Chao Li, Lin Yuan,
  Guang Yang, and Chenghua Lin. 2025.
\newblock \href {https://doi.org/10.18653/v1/2025.findings-emnlp.1100}
  {{Crafting Customisable Characters with LLMs: A Persona-Driven Role-Playing
  Agent Framework}}.
\newblock In \emph{Findings of the Association for Computational Linguistics:
  EMNLP 2025}, pages 20216--20240, Suzhou, China. Association for Computational
  Linguistics.

\bibitem[{Yu and Yu(2021)}]{yu2019midas}
Dian Yu and Zhou Yu. 2021.
\newblock \href {https://doi.org/10.18653/v1/2021.eacl-main.94} {{MIDAS: A
  Dialog Act Annotation Scheme for Open Domain Human--Machine Spoken
  Conversations}}.
\newblock In \emph{Proceedings of the 16th Conference of the European Chapter
  of the Association for Computational Linguistics: Main Volume}, pages
  1103--1120, Online. Association for Computational Linguistics.

\bibitem[{Zang et~al.(2020)Zang, Rastogi, Sunkara, Gupta, Zhang, and
  Chen}]{zang2020multiwoz}
Xiaoxue Zang, Abhinav Rastogi, Srinivas Sunkara, Raghav Gupta, Jianguo Zhang,
  and Jindong Chen. 2020.
\newblock \href {https://doi.org/10.18653/v1/2020.nlp4convai-1.13} {{MultiWOZ
  2.2: A Dialogue Dataset with Additional Annotation Corrections and State
  Tracking Baselines}}.
\newblock In \emph{Proceedings of the 2nd Workshop on Natural Language
  Processing for Conversational AI}, pages 109--117, Online. Association for
  Computational Linguistics.

\bibitem[{Zhang et~al.(2018)Zhang, Dinan, Urbanek, Szlam, Kiela, and
  Weston}]{zhang2018personalizing}
Saizheng Zhang, Emily Dinan, Jack Urbanek, Arthur Szlam, Douwe Kiela, and Jason
  Weston. 2018.
\newblock \href {https://doi.org/10.18653/v1/P18-1205} {{Personalizing Dialogue
  Agents: I have a dog, do you have pets too?}}
\newblock In \emph{Proceedings of the 56th Annual Meeting of the Association
  for Computational Linguistics (Volume 1: Long Papers)}, pages 2204--2213,
  Melbourne, Australia. Association for Computational Linguistics.

\bibitem[{Zhao et~al.(2025)Zhao, Vania, Kayal, Khan, Cohen, and
  Yilmaz}]{zhao2025personalens}
Zheng Zhao, Clara Vania, Subhradeep Kayal, Naila Khan, Shay~B. Cohen, and Emine
  Yilmaz. 2025.
\newblock \href {https://doi.org/10.18653/v1/2025.findings-acl.927}
  {{PersonaLens: A Benchmark for Personalization Evaluation in Conversational
  AI Assistants}}.
\newblock In \emph{Findings of the Association for Computational Linguistics:
  ACL 2025}, pages 18023--18055, Vienna, Austria. Association for Computational
  Linguistics.

\bibitem[{Zheng et~al.(2023)Zheng, Chiang, Sheng, Zhuang, Wu, Zhuang, Lin, Li,
  Li, Xing, Zhang, Gonzalez, and Stoica}]{zheng2024judging}
Lianmin Zheng, Wei-Lin Chiang, Ying Sheng, Siyuan Zhuang, Zhanghao Wu, Yonghao
  Zhuang, Zi~Lin, Zhuohan Li, Dacheng Li, Eric~P. Xing, Haotong Zhang,
  Joseph~E. Gonzalez, and Ion Stoica. 2023.
\newblock {Judging LLM-as-a-Judge with MT-Bench and Chatbot Arena}.
\newblock In \emph{Advances in Neural Information Processing Systems},
  volume~36, pages 46595--46623.

\end{thebibliography}
